\documentclass{article}

\usepackage{graphicx,float}
\usepackage[space]{grffile}
\usepackage{latexsym}
\usepackage{textcomp}
\usepackage{longtable}
\usepackage{tabulary}
\usepackage{booktabs,array,multirow}
\usepackage{amsfonts,amsmath,amssymb}
\usepackage[round]{natbib}
\usepackage{url}
\usepackage{hyperref}
\hypersetup{colorlinks=false,pdfborder={0 0 0}}
\usepackage{etoolbox}
\usepackage{algorithm}
\usepackage{algpseudocode}

\usepackage{authblk}
\usepackage[left=1in,right=1in,top=1in,bottom=1in]{geometry}

\newif\iflatexml\latexmlfalse

\AtBeginDocument{\DeclareGraphicsExtensions{.pdf,.PDF,.eps,.EPS,.png,.PNG,.tif,.TIF,.jpg,.JPG,.jpeg,.JPEG}}

\usepackage[utf8]{inputenc}
\usepackage[english]{babel}

\usepackage{siunitx}
\usepackage{lineno}

\newcommand{\Dmu}[1]{\mathbf{D}\mu_{\encoder_{#1}}}

\title{Ensemble Kalman filter in latent space using a variational autoencoder pair}

\author[1]{Ivo Pasmans}
\author[1]{Yumeng Chen}
\author[2]{Tobias Sebastian Finn}
\author[2]{Marc Bocquet}
\author[1,3]{Alberto Carrassi}

\affil[1]{Department of Meteorology, University of Reading, Reading, UK}
\affil[2]{CEREA, ENPC, EDF R\&D, Institut Polytechnique de Paris, Île-de-France, France}
\affil[3]{Department of Physics ``Augusto Righi'', Universit\'{a} di Bologna, Bologna, Italy}

\newcommand{\reftab}[1]{Table~#1}
\newcommand{\reffig}[1]{Figure~#1}
\newcommand{\refsec}[1]{Section~#1}
\newcommand{\refapp}[1]{Appendix~#1}

\newcommand{\refFig}[1]{Figure~#1}
\newcommand{\refSec}[1]{Section~#1}
\newcommand{\refeq}[1]{Equation~(#1)}
\newcommand{\refEq}[1]{Equation~(#1)}
\newcommand{\refalg}[1]{Algorithm~#1}
\newcommand{\etkfVae}[2]{ETKF-VAE\rlap{\textsuperscript{#1}}\textsubscript{#2}}

\newcommand{\config}[1]{{\it #1}}

\DeclareMathOperator{\Ima}{Im} 
\newcommand{\realSpace}[1]{\mathbb{R}^{#1}}
\newcommand{\modelSpace}{\mathcal{X}}
\newcommand{\latentSpace}{\mathcal{Z}}
\newcommand{\obsSpace}{\mathcal{Y}}
\DeclareMathOperator{\stateSpaceDim}{N_{\modelSpace}}
\DeclareMathOperator{\latentSpaceDim}{N_{\latentSpace}}
\DeclareMathOperator{\ensSize}{M}

\newcommand{\for}[1]{{#1}^{\mathrm{f}}}
\newcommand{\ana}[1]{{#1}^{\mathrm{a}}}
\newcommand{\clima}[1]{{#1}^{\mathrm{clima}}}
\newcommand{\truth}[1]{{#1}^{\mathrm{truth}}}

\newcommand{\ensMean}{\mu_{\ensState}}
\newcommand{\ensCov}{\mathbf{P}_{\ensState}}
\newcommand{\crossCov}{\mathbf{Y}}

\renewcommand{\vec}[1]{\mathbf{#1}}
\newcommand{\ensState}{\vec{X}}
\newcommand{\vecState}{\vec{x}}
\newcommand{\ensInno}{\vec{D}}
\newcommand{\vecInno}{\vec{d}}
\newcommand{\vecObs}{\vec{y}}
\newcommand{\ensLatentState}{\vec{Z}}
\newcommand{\vecLatentState}{\vec{z}}
\newcommand{\ensLatentInno}{\vec{F}}
\newcommand{\vecLatentInno}{\vec{f}}
\newcommand{\obsOp}{H}
\newcommand{\obsOpLin}{\mathbf{H}}
\newcommand{\trace}{\mathrm{tr}}
\newcommand{\T}{\mathrm{T}}
\newcommand{\obsCov}{\mathbf{R}}
\newcommand{\anomaly}[1]{\tilde{#1}}

\newcommand{\dx}[1][x]{\,\mathrm{d}{#1}}
\DeclareMathOperator{\heaviside}{\mathcal{H}}

\DeclareMathOperator{\Pnorm}{\mathcal{N}}

\DeclareMathOperator{\Pskew}{\mathcal{S}}
\DeclareMathOperator{\Puniform}{\mathcal{U}}
\newcommand{\encoder}{\phi}
\newcommand{\decoder}{\theta}
\DeclareMathOperator{\expectation}{\mathbb{E}}
\DeclareMathOperator{\KL}{\mathrm{KL}}
\DeclareMathOperator{\ELBO}{\mathcal{L}}
\newcommand{\vaeMean}{\mu}
\newcommand{\vaeCov}{\Sigma}

\newcommand{\nextsimdg}{$\mathrm{neXtSIM_{DG}}\,$}

\begin{document}

\maketitle
\begin{abstract}
Popular (ensemble) Kalman filter data assimilation (DA) approaches assume that the errors in both the a priori estimate of the state and those in the observations are Gaussian. For constrained variables, e.g. sea ice concentration or stress, such an assumption does not hold. The variational autoencoder (VAE) is a machine learning (ML) technique that allows to map an arbitrary distribution to/from a latent space in which the distribution is supposedly closer to a Gaussian. We propose a novel hybrid DA-ML approach in which VAEs are incorporated in the DA procedure. Specifically, we introduce a variant of the popular ensemble transform Kalman filter (ETKF) in which the analysis is applied in the latent space of a single VAE or a pair of VAEs. In twin experiments with a simple circular model, whereby the circle represents an underlying submanifold to be respected, we find that the use of a VAE ensures that a posteri ensemble members lie close to the manifold containing the truth. Furthermore, online updating of the VAE is necessary and achievable when this manifold varies in time, i.e. when it is non-stationary. We demonstrate that introducing an additional second latent space for the observational innovations improves robustness against detrimental effects of non-Gaussianity and bias in the observational errors but it slightly lessens the performance if observational errors are strictly Gaussian.

\end{abstract} 

\section{Introduction}
\label{sec:intro}

Data assimilation (DA) aims to provide a more precise estimate of the true state of a system by combining a prior guess in the form of a probability distribution with observations \cite{carrassi_data_2018}. Data assimilation is widely used in operational atmosphere, ocean and sea ice forecasting \citep{waters_implementing_2015,de_rosnay_coupled_2022,qin_global-ocean-data_2023,inverarity_met_2023,buehner_modular_2024} and will be used as well by the \nextsimdg sea ice model \citep{richter_dynamical_2023,jendersie_gpu-parallelization_2024} that is currently being developed as part of the Scale Aware Sea Ice Project (SASIP) (\url{https://sasip-climate.github.io/}). In previous studies, we have explored possibilities of tailoring DA to the discontinuous Galerkin numerical core used by \nextsimdg \citep{pasmans_tailoring_2024} and to infer sea ice parameters using an ensemble Kalman filter (EnKF) \citep{chen_multivariate_2024}. Though intentionally fully conceptual in its nature, this work is also motivated by the challenges posed to DA by the complex physics of the sea ice, in particular the presence of nonlinear relations and constraints in the sea ice dynamics. These issues are, however, not exclusive to sea ice modeling, but instead are pervasive in many other branches of the climate and weather prediction at large; see e.g. the modelling of humidity in atmospheric models. 

The EnKF \citep{evensen_sequential_1994} is one of the most popular DA methodologies. In the EnKF the probability distribution for the true state is assumed to be a Gaussian with mean and covariance estimated from an ensemble of model runs. During the update step a correction is added to the ensemble members by taking into account the information from the observations. Following this, the ensemble members are propagated to the next time step with a dynamical model. In real scenarios, the use of the EnKF can be challenging. The computational demands of geophysical models can easily make running large ensembles prohibitive. The use of smaller, computationally more affordable, ensembles introduces sampling errors that need to be mitigated. Remediation of the effect of these errors requires the application of additional techniques such as ensemble inflation \citep{ehrendorfer_review_2007,whitaker_evaluating_2012} and localisation \citep{ehrendorfer_review_2007,morzfeld_theory_2022}. Third, the EnKF assumes that the errors in the prior estimate, background errors, and observations are unbiased and Gaussian. Such assumptions generally do not hold. Finally, EnKF is derived by the linear estimation theory, i.e. apart from (structured) noise there is a linear relationship between observed differences between observations and model predictions (a.k.a the innovation) on one hand and the DA correction on the other hand. This implies that there is an affine space, consisting of elements that can be decomposed as the initial guess plus a possible DA correction, in which all elements are model solutions as well. Although such spaces of possible solutions exist for linear models, their existence is not guaranteed for nonlinear models. This implies that for nonlinear models the DA process might produce physically non-realisable states. These assumptions are especially problematic when using sea ice models with a brittle rheology \citep{dansereau_maxwell_2016,olason_new_2022}, such as \nextsimdg. These models contain strong nonlinear relationships between sea ice damage and sea ice viscosity and elasticity. Furthermore, errors and physically realiseable sea ice states are constrained by several bounds. While some of these are simple, e.g. sea ice concentration must lie between $0$ and $1$, others bounds imposed on sea ice stresses by the Mohr-Coulomb relation are a nonlinear function of the sea ice state themselves. 

\subsection{DA in the latent space}

Over the last years there has been a proliferation of works fusing DA with machine learning (ML). Some exemplary studies use ML-based DA to learn model dynamics or model error corrections \citep{bocquet_bayesian_2020,bocquet_online_2020,brajard_combining_2020,brajard_combining_2021-1,arcucci_deep_2021,farchi_using_2021-1,farchi_online_2023}, or DA itself \citep{mccabe_learning_2021,boudier_data_2023,bocquet_accurate_2024,chinellato_state_2024}. In other cases ML is blended with DA in an attempt to address some of the aforementioned challenges of DA: the computational cost, the need for inflation and localisation, physical imbalances and the violation of the Gaussian or the quasi-linear assumptions. In the following, we provide a very short and inevitably non-exhaustive overview of these attempts; they are schematically reported in \reftab{\ref{tab:literature}}. Our scope is to provide essential elements of the context within which our current study is rooted. Hence we focus primarily on studies in which the issues are addressed with the aid of a second, often lower dimensional, space, called the latent space. The reader may find more extensive review in \citet{buizza_data_2022-2,cheng_machine_2023-2,bach_inverse_2024,shlezinger_ai-aided_2024}.

\begin{table}[H]
\centering
\normalsize
\begin{tabulary}{\textwidth}{CCCCC}
    {\bf Study} &  {\bf ML methods} &  {\bf Forward propagation} & {\bf DA method} & {\bf Observation operator} \\
    \hline
    \citet{canchumuni_towards_2019} & VAE & identity & latent ensemble smoother & default \\ 
    \citet{mack_attention-based_2020} & autoencoder & physical model & latent 3DVar & Penrose inverse, decoder, identity on the latent space \\ 
    \citet{amendola_data_2021} & CNN, LSTM & latent model & latent EnKF & embedding \\
    \citet{grooms_analog_2021} & VAE & none & physical EnOI & default \\
    \citet{peyron_latent_2021}  & autoencoder, residual network  & latent ensemble & latent ETKF & decoder, default \\
    \citet{bao_variational_2022} & VAE &
    identity & latent EnKF & decoder, default \\
    \citet{cheng_data-driven_2022} & autoencoder, LSTM & latent model & latent 3DVAR &
    identity in latent space  \\
    \citet{maulik_efficient_2022} & LSTM, POD & principal component model & 4DVAR & Penrose inverse - default \\
    \citet{pawar_equation-free_2022} & LSTM & principal component model & DEnKF & Penrose inverse - identity latent space\\
     \citet{rozet_score-based_2023} & diffusion model & physical model & deep Kalman filter & default  \\ 
    \citet{finn_representation_2024} & diffusion model & physical model & ETKF & default \\
    \citet{huang_diffda_2024} & diffusion model & physical model& latent state nudging & embedding \\
    \citet{luk_learning_2024} & linear transformation & physical model & physical & default \\
    \citet{melinc_3d-var_2024} & VAE & physical model & latent 3DVAR & decoder mean - default \\
    \citet{qu_deep_2024}  & diffusion model & physical & score function nudging & default \\
    \citet{si_latent-ensf_2024} & VAE, diffusion model & physical model & score function nudging & identity on latent space 
\end{tabulary}
\caption{Summary of various ML-DA studies based leveraging on the small dimension latent space. The ML algorithms used, the way states are propagated forward in time, the DA method used and the space to which it is applied and additional information on how the observation operator that acts on a physical state (default operator) is modified to act on the latent space. Here VAE refers to variational autoencoder, EnKF to ensemble Kalman filter and ETKF to ensemble transform Kalman filter. \label{tab:literature}}
\end{table}

One aim of the ML-for-DA schemes is to reduce the computational burden of DA. For instance, \citet{maulik_efficient_2022} reduces the computational cost of the forecast model by 1) reducing the dimensionality of the model states with the aid of a principal orthogonal decomposition (POD), and, 2) training a recurrent neural network (RNN) to predict those coefficients. These coefficients are then corrected by 4DVAR \cite{carrassi_data_2018}, a variational DA method, exploiting automatic differentiation of the RNN. A similar concept is followed by \citet{amendola_data_2021,peyron_latent_2021,akbari_blending_2023} but they replace the POD with either convolutional neural networks (CNN) or an autoencoder that map the states to/from a low-dimensional latent space where the DA correction takes place. They propagate an ensemble of model trajectories in this latent space and apply an EnKF to it. Yet, this approach introduces another source of nonlinearity, since the observation operator becomes the composition of the operator on the original state space with the generally nonlinear decoder. To avoid or mitigate this problem, \citet{cheng_data-driven_2022} introduces a second autoencoder for observations while  \citet{pawar_equation-free_2022} replaces the autoencoder with a linear projection on the principal components. 

Machine learning schemes capable of cheaply generating large ensembles have also been proposed to eliminate the need for inflation and localisation. One such scheme is the variational autoencoder \citep[VAE, ][]{kingma_introduction_2019}. Like the autoencoder, the VAE consists of an encoder-decoder pair, but now these functions output probability distributions instead of a single state. Examples of this approach can be found in \citet{grooms_analog_2021} in which ensemble members are drawn from probability distributions produced by applying the encoder-decoder to the a priori state. A similar approach, but with a denoising diffusion model instead of a VAE can be found in \citet{finn_representation_2024}. In contrast to \citet{grooms_analog_2021}, \citet{melinc_3d-var_2024} applied DA in the latent space of a VAE. The posterior distribution in the latent space is then mapped by the decoder back to the physical space. As the decoder has been trained to reproduce the climatological distribution of atmospheric states, the mapped physical state are expected to respect physical relations in the atmosphere. As the probability distribution depends on the a priori state, the generated ensemble will vary in time. However, the spread of the ensemble in latent space has to be specified by the data assimilator and hence the uncertainty in the prior and posterior state might not converge to its true value over time. 

The presence of non-Gaussian errors in DA has traditionally been addressed by transforming realisations of the non-Gaussian distribution to realisations of a Gaussian distribution using anamorphosis after which a conventional DA method can be applied. One common application of Gaussian anamorphosis is the transformation of strictly positive variables, such as sea ice concentration, from a log-normal to a normal distribution \citep{polavarapu_data_2005,fletcher_hybrid_2006,bocquet_beyond_2010,simon_gaussian_2012,song_incremental_2012}. Gaussian distributions can also be constructed from arbitrary distributions using quantile matching \citep{bertino_sequential_2003,beal_characterization_2010,simon_gaussian_2012,metref_non-gaussian_2014,kotsuki_assimilating_2017,grooms_comparison_2022}. The method is however not practically applicable in all circumstances. The flexible quantile matching method requires the availability of histograms of the background errors to infer the random variable distribution. Construction of such histograms necessitates the ergodic error assumption or large ensembles, in the case of e.g. the rank regression Kalman filter \citep{anderson_non-gaussian_2010,anderson_nonlinear_2019}. The former limits the amount of spatial variability of the distribution that can be captured by the transformation, and the latter is computationally costly. In high-dimensional applications, the quantile matching is only practical for univariate variables. A ML alternative to quantile matching is provided by normalizing flows \citep{tabak_family_2013}. In these flows the relation between the arbitrary probability distribution and the standard normal is constructed by neural networks performing a sequence of coordinate transformations. However, this approach could face its own challenges in high-dimensional applications due to the need for multiple determinants computations.

An alternative way to deal with non-Gaussianity is by applying DA in the latent space of a diffusion model. This is an approach followed by \citet{amendola_data_2021,rozet_score-based_2023,huang_diffda_2024,qu_deep_2024,si_latent-ensf_2024}. These models can deal with arbitrary background and observation error distributions, but conditioning the generated model output on the observations is, however, non-trivial. Furthermore, the reverse denoising algorithm used to generate the corrected states can suffer from numerical instabilities \citep{qu_deep_2024}. Consequently, DA with diffusion models will not be pursued in this work. 

\subsection{The double ETKF-VAE}

In this work, we propose applying the ensemble transform Kalman filter (ETKF), a flavour of DA, in the latent space of a VAE. One of the reasons for doing so is that we want to address the concern around non-Gaussianity mentioned before. The variational autoencoder is trained to relate an arbitrary distribution to a standard normal distribution. Although the relationship is not perfect, we expect that the distribution of the ensemble members in the latent space ends up to be closer to normal than the one in physical space. Hence, when the ETKF is applied in the latent space, the Gaussian assumptions under which the ETKF solution is optimal are closer to be satisfied. Consequently, we hypothesize that this ETKF-VAE setup will outperform a conventional ETKF. 

As shown in the last column of \reftab{\ref{tab:literature}}, the definition of the observation operator and observational error covariance is non-trivial in DA when using a latent space. In this work, we will also present our solution to this problem: a second VAE. The objective of this second VAE is twofold. First, it should bring the distribution of the ensemble members projected into the space of observations to be more Gaussian. Second, it should remove any non-Gaussianity in the observational errors. As such, we expect that, certainly when the observational errors are non-Gaussian, the double ETKF-VAE would outperform the setup in which the VAE is applied solely to the ensemble members. 

Finally, we expect that by mapping the corrected ensemble members from latent space back to physical space using the VAE, the post-DA ensemble members remain physically consistent. In this aspect, we are building on the work by \citet{canchumuni_towards_2019,bao_variational_2022} who also apply DA in the latent space of a VAE. However, we expand on their work in two directions. First, by adding the aforementioned second VAE. Second, we will be using a cycling setup, propagating the DA correction forward in time, while \citet{canchumuni_towards_2019,bao_variational_2022} try to find a correction for a single time only. 

The outline of the paper is as follows. In \refsec{\ref{sec:theory}} the mathematics behind the VAE are explained, our novel double VAE-DA method is introduced and the necessary modifications to the classical EnKF algorithm are discussed. \refSec{\ref{sec:setup}} contains the description of the ideal test-ground model and of the VAE architecture used in the experiments. The experiments themselves and their outcomes can be found in \refsec{\ref{sec:experiments}} while the discussion of the results together with the conclusion are drawn in \refsec{\ref{sec:conclusions}}. 

\section{Ensemble Kalman filter-variational autoencoder hybrids}
\label{sec:theory} 

The hybrid DA-ML scheme introduced in this study is based on the cornerstone idea of applying a modified ensemble transform Kalman filter (ETKF) in the latent space. In this section, we will first introduce the VAE in a general setting and then explain how it can be combined with the ETKF.

\subsection{Variational autoencoder}
\label{sec:vae}

The VAE is a type of generative ML technique consisting of an encoder and a decoder. Let $\{\vecState_{1},\ldots,\vecState_{\ensSize}\} \subset \modelSpace \subseteq \realSpace{\stateSpaceDim}$ be a set of realisations from an unknown probability distribution. In this section no assumptions are imposed on the probability distribution, but from \refsec{\ref{sec:vae-da}} onward it is assumed that the realisations will be either members of the forecast ensemble or members of the ensemble of innovations at the time at which a DA correction is calculated. The probability density function (PDF) for this distribution can be expressed by explanatory latent variables $\vecLatentState \in \realSpace{\latentSpaceDim}$ with 

\begin{equation*}
	p_{\modelSpace}(\vecState) = \int p(\vecState| \vecLatentState)p_{\latentSpace}(\vecLatentState)  \dx[\vecLatentState].
\end{equation*}

Here, $p_{\latentSpace}(\vecLatentState) \overset{\mathrm{def}}{=}\Pnorm(\vecLatentState;\vec{0}_{\latentSpace},\vec{I}_{\latentSpace})$ is the PDF of the standard normal distribution with $\vec{0}_{\latentSpace}$ and $\vec{I}_{\latentSpace}$ being the zero vector and the identity matrix in $\realSpace{\latentSpaceDim}$ respectively. The decoder, $p_{\decoder}(\vecState|\vecLatentState)$, provides the PDF of the transformation from a given latent state $\vecLatentState$ to a state $\vecState$ in $\modelSpace$. Similarly, the encoder, $q_{\encoder}(\vecLatentState|\vecState)$ provides a PDF for the transformation of a state $\vecState$ to a latent state $\vecLatentState$.  The VAE aims to find parameters such that the parametrised PDF $p_{\decoder}(\vecState)$ approximates as good as possible the PDF $p_{\modelSpace}$ by minimising the Kullback-Leibner (KL) divergence $\KL\left[p_{\modelSpace}(\vecState) ||  p_{\decoder}(\vecState)\right]$ \citep{rezende_stochastic_2014,kingma_introduction_2019}. The KL divergence is a positive measure that obtains its minimum of $0$ if and only if $p_{\modelSpace}=p_{\decoder}$ nearly everywhere. Expanding KL divergence and introducing the decoder gives

\begin{subequations}
\begin{align}
\KL\left[p_{\modelSpace}(\vecState) ||  p_{\decoder}(\vecState)\right] 
\overset{\mathrm{def}}{=} &\int p_{\modelSpace}(\vecState) \ln \frac{p_{\modelSpace}(\vecState)}{p_{\decoder}(\vecState)} \dx[\vecState]  = 
\int p_{\modelSpace}(\vecState) \ln p_{\modelSpace}(\vecState) \dx[\vecState] \nonumber \label{eq:minfunction:a}\\
&+ \int p_{\modelSpace}(\vecState) q_{\encoder}(\vecLatentState|\vecState) \ln \frac{q_{\encoder}(\vecLatentState|\vecState) p_{\decoder}(\vecLatentState|\vecState)}{p_{\decoder}(\vecState)q_{\encoder}(\vecLatentState|\vecState) p_{\decoder}(\vecLatentState|\vecState)} \dx[\vecLatentState] \dx[\vecState], \\
=& \int p_{\modelSpace}(\vecState) \ln p_{\modelSpace}(\vecState) \dx[\vecState]
- \int p_{\modelSpace}(\vecState) \KL[q_{\encoder}(\vecLatentState|\vecState)||p_{\decoder}(\vecLatentState|\vecState)] \dx[\vecState]  
- \int p_{\modelSpace}(\vecState) \ELBO(\encoder,\decoder,\vecState) \dx[\vecState],\label{eq:minfunction}
\end{align}
with $q_{\encoder}(\vecLatentState|\vecState)$, the encoder defined as a parametrised PDF, and
\begin{equation} \ELBO(\encoder,\decoder,\vecState) \overset{\mathrm{def}}{=} \int q_{\encoder}(\vecLatentState|\vecState) \ln \frac{p_{\theta}(\vecState,\vecLatentState)}{q_{\encoder}(\vecLatentState|\vecState)} \dx[\vecLatentState] = \int q_{\encoder}(\vecLatentState|\vecState) \ln p_{\decoder}(\vecState|\vecLatentState) \dx[\vecLatentState] - \int q_{\encoder}(\vecLatentState|\vecState)\ln \frac{q_{\encoder}(\vecLatentState|\vecState)}{p_{\decoder}(\vecLatentState)} \dx[\vecLatentState],
\label{eq:elbo}
\end{equation}
\end{subequations}
being the evidence lower bound (ELBO). 
In the second line of \refeq{\ref{eq:minfunction:a}} we made use of the relation 
\begin{equation}
\ln \frac{1}{p_{\decoder}(x)} = \ln \frac{1}{p_{\decoder}(x)} \int q_{\encoder}(\vecLatentState | \vecState)  \dx[\vecLatentState]  = \int q_{\encoder}(\vecLatentState | \vecState) \ln \frac{1}{p_{\decoder}(\vecState)} \dx[\vecLatentState].
\end{equation}
The weights of the VAE $\encoder$ and  $\decoder$ are found by maximising the expectation value of the ELBO (last term in \refeq{\ref{eq:minfunction}}). Loosely stated, the maximisation of the expectation value of the ELBO will result in a decoder that approximately minimises the left-hand side of \refeq{\ref{eq:minfunction}} while simultaneously producing an encoder $p_{\encoder}(\vecLatentState|\vecState)$ that is an approximate ``inverse'' of the decoder $p_{\decoder}(\vecState|\vecLatentState)$. Alternatively, one can view maximisation of the ELBO as minimisation of the reconstruction error, $p_{\decoder}(\vecState | \vecLatentState)$, in the first term on the right-hand side of \refeq{\ref{eq:elbo}} regularised by the need for the learned encoder, $q_{\encoder}(\vecLatentState | \vecState)$, to stay close to the the standard normal, $p_{\decoder}(\vecLatentState)$. 

As is conventional, the encoder and decoder distributions are assumed to be Gaussian with diagonal covariances. That is to say $p_{\encoder}(\vecLatentState|\vecState)=\Pnorm(\vecLatentState;\vaeMean_{\encoder}(\vecState),\vaeCov_{\encoder}(\vecState))$ and $p_{\decoder}(\vecState|\vecLatentState)=\Pnorm(\vecState;\vaeMean_{\decoder}(\vecLatentState),\vaeCov_{\decoder}(\vecLatentState))$ in which the $\vaeMean_{\encoder}$, $\vaeMean_{\decoder}$, $\ln \vaeCov_{\encoder}$ and $\ln \vaeCov_{\decoder}$ are provided as outputs from neural networks. The exact architectures of these networks as used in this study are given in \refsec{\ref{sec:experiments}}. When the expectation value of the ELBO (i.e. the last term in \refeq{\ref{eq:minfunction}}) is replaced by its single-sample Monte-Carlo approximation it can be written as:

\begin{subequations}
\begin{align}
    \vecLatentState_{\encoder} \sim& \Pnorm(\vaeMean_{\encoder}(\vecState),\vaeCov_{\encoder}(\vecState)) \label{eq:sample}, \\
	\ln \vaeCov'_{\decoder} \overset{\mathrm{def}}{=} & (1-\gamma)\ln \vaeCov_{\decoder}(\vecLatentState_{\encoder}) + \gamma \ln \vaeCov_{\mathrm{def}} \label{eq:regularization},\\
	2\int p_{\modelSpace}(\vecState) q_{\encoder}(\vecLatentState|\vecState) \ln p_{\decoder}(\vecState|\vecLatentState) \dx[\vecLatentState] \dx[\vecState] \approx& -\ln \det(\vaeCov'_{\decoder}) - ||\vaeCov_{\decoder}'^{-\frac{1}{2}}(\vecState-\mu_{\decoder}(\vecLatentState_{\encoder}))||^{2}
	-||\ln \vaeCov'_{\decoder} - \ln \vaeCov_{\decoder}(\vecLatentState_{\encoder})||^2 ,\label{eq:reconstruction}\\ 
	2\int p_{\modelSpace}(\vecState)q_{\encoder}(\vecLatentState|\vecState)\ln \frac{q_{\encoder}(\vecLatentState|\vecState)}{p_{\decoder}(\vecLatentState)} \dx[\vecLatentState] \dx[\vecState] \approx&  ||\vaeMean_{\encoder}(\vecState)||^2+
	\trace(\vaeCov_{\encoder}(\vecState)) - \ln \det(\vaeCov_{\encoder}(\vecState)) - \latentSpaceDim, \label{eq:kl} 
\end{align}
\end{subequations}
with the evaluation of $\vecLatentState$-integral in \refeq{\ref{eq:kl}} taken from \citet{zhang_properties_2023}, \refeq{\ref{eq:sample}} indicating that $\vecLatentState_{\encoder}$ is drawn from a normal distribution with mean $\mu_\encoder(\vecState)$ and covariance $\vaeCov_{\encoder}(\vecState)$,  $\gamma$ an epoch-dependent regularisation factor that goes to zero as the number of epochs goes to infinity and $\vaeCov_{\mathrm{def}}$ a reference covariance to be specified. The $\gamma$-regularization is there to avoid well-documented convergence issues with the ELBO \citep{rezende_taming_2018-1,dai_diagnosing_2019-1} without having to resort to fixing the decoder variance $\vaeCov_{\decoder}$.

\subsection{Ensemble transform Kalman filter revisited}
\label{sec:etkf}

The Kalman filter (KF) is the analytic complete solution of the sequential Bayesian filter problem, under the assumption of Gaussian errors and linear dynamical and observation models. The KF estimates the unknown Gaussian PDF of the true state of the system of interest $\truth{\vecState} \sim \Pnorm(\ensMean,\ensCov)$ with $\ensMean$ and $\ensCov$ being a function of both time and assimilated noisy observations $\vecObs_{t} \sim \Pnorm(\obsOp(\truth{\vecState}_{t}), \obsCov)$, with $\obsOp=\obsOpLin$ the linear observation operator. The posterior mean and covariance are given by \citep[section 6.1]{evensen_data_2022}:

\begin{subequations}
 \label{eq:kf}
\begin{align}
	\ana{\ensMean} =& \for{\ensMean}+\for{\ensCov}\obsOpLin^{\T} (\obsOpLin \for{\ensCov} \obsOpLin^{\T}+\obsCov)^{-1} (\vecObs-\obsOpLin \for{\ensMean})\label{eq:anaMean},  \\
	  \ana{\ensCov} =& \for{\ensCov}-\for{\ensCov}\obsOpLin^\T (\obsOpLin \for{\ensCov} \obsOpLin^\T + \obsCov)^{-1}\obsOpLin \for{\ensCov}. \label{eq:anaCov}
\end{align}
\end{subequations}

The ensemble Kalman filter (EnKF) relaxes the requirement that the dynamical model and observation operator have to be linear and estimates the mean $\ensMean$ and covariance $\ensCov$ from an ensemble of, possibly nonlinear, member runs. I.e. $\ensMean=\frac{1}{\ensSize}\ensState \vec{1}_{\ensSize}$, $\ensCov=\frac{1}{\ensSize-1}\anomaly{\ensState}\anomaly{\ensState}^\T$ and $\crossCov=\obsOpLin \ensState$. Here $\anomaly{\cdot}$ indicates that the ensemble mean has been removed from each column to form the anomaly matrices, i.e. $\anomaly{\ensState}=\ensState(\vec{I}-\frac{1}{\ensSize}\vec{1}_{\ensSize}\vec{1}_{\ensSize}^\T)$ and $\vec{1}_{\ensSize}\in \realSpace{\ensSize}$ a vector having ones as its elements.  With these substitutions \refeq{\ref{eq:kf}} can be rewritten as

\begin{subequations}
\begin{align}
    \ana{\ensState} =& \for{\ensState} + \for{\anomaly{\ensState}} \crossCov^\T \vec{U} \vec{S}^{-1} \vec{U}^\T\vec(\vec{y}-\frac{1}{M}\obsOpLin\for{\ensState}\vec{1}_{\ensSize})\vec{1}^\T_{\ensSize} + \for{\anomaly{\ensState}}\crossCov^\T\vec{U}(\vec{I}-\vec{S}^{-1})^{1/2},   \label{eq:etkf} \\
    \frac{1}{\ensSize-1}\anomaly{\crossCov} \anomaly{\crossCov}^\T + \obsCov \overset{\mathrm{SVD}}{=}& \vec{U} \vec{S} \vec{U}^\T,
\end{align}
\end{subequations}

with the last line of \refeq{\ref{eq:etkf}} defining a singular value decomposition (SVD).

In anticipation of the work in \refsec{\ref{sec:vae-da}}, we slightly modify the ETKF in \refeq{\ref{eq:etkf}} using $\obsOpLin \for{\ensCov} \obsOpLin^\T + \obsCov \approx \frac{1}{K-1}\anomaly{\ensInno}_{K}\anomaly{\ensInno}_{K}^\T$ from \citet{desroziers_diagnosis_2001} with $\ensInno_{K} \in \realSpace{N_{\vecObs} \times K}$ a matrix having perturbed innovations as ensemble members, i.e. the $k$th column of $\ensInno$ is given by $\vecObs+\epsilon^{\vecObs}_{k}-\obsOpLin \vecState_{m_{k}}$ with $K\gg\ensSize$, $N_{\obsSpace}$ the number of observations, $\epsilon^{\vecObs}_{k}$ is a realisation of the observational error distribution and each $m_{k}$ is randomly chosen from $\{1,2,\ldots,\ensSize\}$. Simultaneously, we rewrite $\obsOpLin \for{\ensCov} = \frac{1}{\ensSize-1}\crossCov \for{\anomaly{\ensState}} = - \anomaly{\ensInno}_{\ensSize}\for{\anomaly{X}}$ with $\ensInno = \vec{y} \vec{1}_{\ensSize}^\T - \obsOpLin \for{\ensState} \in \realSpace{N_{\obsSpace}\times \ensSize}$, the matrix having the innovation of each ensemble member as columns. After making these substitutions, \refeq{\ref{eq:etkf}} can be rewritten as

\begin{subequations}
\begin{align}
    \ana{\ensState} =& \for{\ensState} - \frac{1}{M} \for{\anomaly{\ensState}} \anomaly{\ensInno}_{\ensSize}^\T \vec{U} \vec{S}^{-1} \vec{U}^\T \ensInno_{\ensSize} \vec{1}_{\ensSize} \vec{1}^\T_{\ensSize} - \for{\anomaly{\ensState}}\anomaly{\ensInno}_{\ensSize}^\T\vec{U}(\vec{I}-\vec{S}^{-1})^{1/2}, \label{eq:etkfD} \\
    \frac{1}{K-1} \anomaly{\ensInno}_{K} \anomaly{\ensInno}_{K} \overset{\mathrm{SVD}}{=}& \vec{U} \vec{S} \vec{U}^\T.
    \end{align}
\end{subequations}

\subsection{Single and double VAE-DA}
\label{sec:vae-da}

In this section, the DA configurations that are going to be used in the experiments of \refsec{\ref{sec:experiments}} are formulated. In particular, we describe how the matrices $\for{\ensState}$, $\ana{\ensState}$, $\ensInno_{K}$ and $\ensInno_{\ensSize}$ as well as the anomaly matrices ($\anomaly{\for{\ensState}}$,$\anomaly{\for{\ensState}}$,$\ldots$) introduced in \refsec{\ref{sec:etkf}} are replaced with equivalents lying in the latent space of one (\config{single ETKF-VAE} configuration) or two VAEs (\config{double ETKF-VAE} configuration).  The description of the architecture and training procedures for these VAEs is postponed to \refsec{\ref{sec:architecture}}. The main work hypothesis at the basis of the proposed approaches is that in the latent space the ensembles, i.e., the column vectors in $\for{\ensState}$, $\ana{\ensState}$, $\ensInno_{K}$ and $\ensInno_{\ensSize}$ respectively are more Gaussian distributed in the latent space than in the state space. 

\subsubsection{Single ETKF-VAE}
\label{sec:single}

\begin{figure}[H]
	\centering
	\includegraphics[width=\textwidth,keepaspectratio=true]{"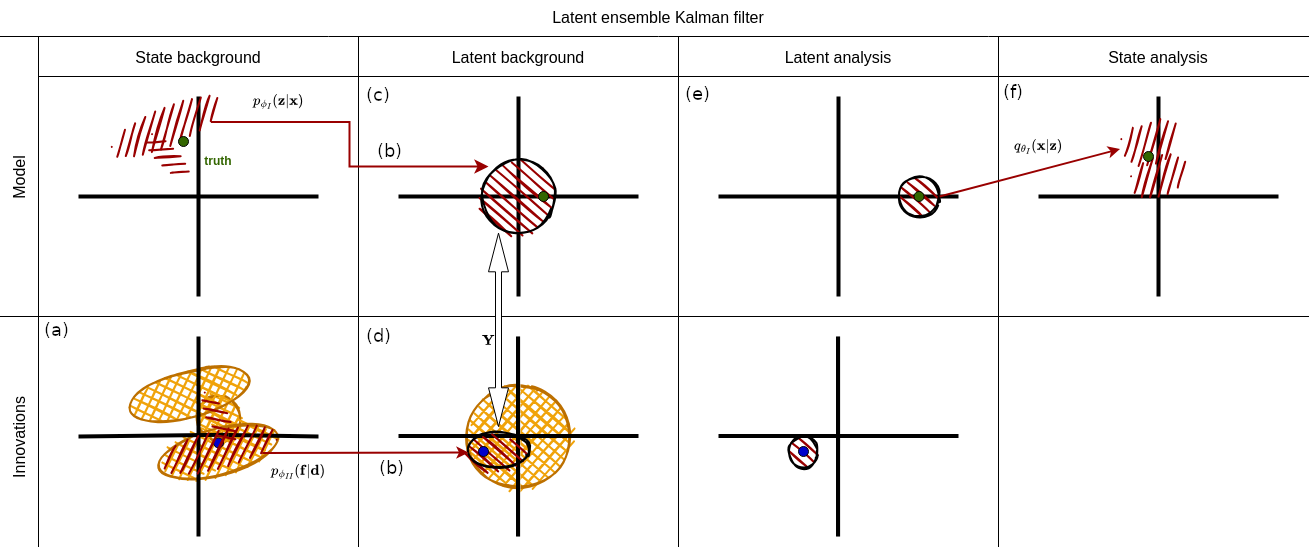"}
	\caption{Schematic overview of the (top) \config{single ETKF-VAE} and (top+bottom) \config{double ETKF-VAE} approach. (a) Alternative innovations are generated by drawing ensemble members and adding realisations of the observational error, (b) the first and second VAE are trained on the forecast ensemble and alternative innovations respectively, (c) the first encoder is used to sample one ensemble member in latent space for each ensemble member in state space, (d) the innovation-encoder is used to sample $K$ perturbed innovations and $\ensSize$ unperturbed innovations in latent space (e) the ETKF is performed using the ensembles of states, perturbed innovations and innovations, (f) for each member in the analysis ensemble the first decoder samples a member in the state space. \label{fig:schematic}}
\end{figure}

In the \config{single ETKF-VAE}, $\for{\ensState}$ and $\ana{\ensState}$ in \refeq{\ref{eq:etkfD}} are replaced by their latent space counterparts, $\for{\ensLatentState} \in \realSpace{N_{\latentSpace_{1}} \times \ensSize}$ and $\ana{\ensLatentState} \in \realSpace{N_{\latentSpace_{1}} \times \ensSize}$ respectively. Mutatis mutandis, the same is happening with $\for{\anomaly{\ensState}}$ and $\ana{\anomaly{\ensState}}$. Their vector columns, the ensemble members in the latent space, $\for{\vecLatentState}_{m}$ and $\ana{\vecLatentState}_{m}$, are related to the original ensemble members in the physical space, the column vectors $\for{\vecState}_{m}$ and $\ana{\vecState}_{m}$, via a VAE trained on model states (see \refsec{\ref{sec:setup}} for the details). Their statistical relations are  $\for{\vecLatentState}_{m} \sim \Pnorm(\vaeMean_{\encoder_{I}}(\for{\vecState}_{m}),\vaeCov_{\encoder_{I}}(\for{\vecState}_{m}))$ and $\ana{\vecState}_{m} \sim \Pnorm(\vaeMean_{\decoder_{I}}(\ana{\vecLatentState}_{m}),\vaeCov_{\decoder_{I}}(\ana{\vecLatentState}_{m}))$. Next to this, the linear observation operator $\obsOpLin$ appearing in \refsec{\ref{sec:etkf}} is replaced with a potentially nonlinear operator $\obsOp$. In the \config{single ETKF-VAE} $\ensInno_{K}$ and $\ensInno_{\ensSize}$ remain as defined in \refsec{\ref{sec:etkf}}. In particular, any non-Gaussianity present in $\ensInno_{K}$ and $\ensInno_{\ensSize}$ stemming from the ensemble contained in $\for{\ensState}$ via $\obsOp$, the nonlinearity of $\obsOp$ or the non-Gaussianity of the observational errors remains unaddressed. This \config{single ETKF-VAE} is illustrated in the first row of \reffig{\ref{fig:schematic}} and in \refalg(\ref{alg:single}) in \refapp{\ref{app:algorithms}}.

\subsubsection{Double ETKF-VAE}
\label{sec:double}

In the \config{double ETKF-VAE} a second VAE is trained on samples of the form $\obsOp(\for{\vecState}_{m_{i}}) + \epsilon^{\vecObs}_{m_{i}} - \obsOp (\for{\vecState}_{m_{j}})$ with $m_{i}$, $m_{j}$ randomly chosen from $\{1,2,\ldots,\ensSize\}$ and $\epsilon_{\vecObs}^{(i)}$ a realisation of the observational error probability distribution under the condition that $\truth{\vecState}=\vecState^{m_{i}}$. These vectors represent possible innovations associated with different ensemble in the absence of additional knowledge about the truth state. This includes any knowledge contained in the observation. 
Notice that had the ensemble been infinite, the columns of $\ensInno_{\ensSize}$ and $\ensInno_{K}$ would be among the innovations created in this way. 

The encoder part of this second VAE is then used to sample for each of the columns in $\ensInno_{\ensSize}$ a vector in the second latent space $\latentSpace_{2}$. These are then collated as columns of $\ensLatentInno_{\ensSize} \in \realSpace{\latentSpace_2 \times \ensSize}$. Similarly, a vector is sampled for each of the columns of $\ensInno_{K}$ creating $\ensLatentInno_{K} \in \realSpace{N_{\latentSpace_2} \times K}$. $\ensLatentInno_{\ensSize}$ and $\ensLatentInno_{K}$ replace $\ensInno_{\ensSize}$ and $\ensInno_{K}$ respectively in \refeq{\ref{eq:etkfD}}.  In addition to this $\for{\ensState}$ and $\ana{\ensState}$ are replaced with $\for{\ensLatentState} \in \realSpace{\latentSpace_1 \times \ensSize}$ and $\ana{\ensLatentState} \in \realSpace{N_{\latentSpace_1} \times \ensSize}$ respectively as was already outlined in \refsec{\ref{sec:single}}. \refEq{\ref{eq:etkfD}} after these substitutions becomes

\begin{subequations}
\begin{align}   
    \ana{\ensLatentState} =& \for{\ensLatentState} - \frac{1}{M} \vec{1}^\T_{\ensSize}\for{\anomaly{\ensLatentState}} \anomaly{\ensLatentInno}_{\ensSize}^\T \vec{U} \vec{S}^{-1} \vec{U}^\T \ensLatentInno_{\ensSize} \vec{1}_{\ensSize} - \for{\anomaly{\ensLatentState}}\anomaly{\ensLatentInno}_{\ensSize}^\T\vec{U}(\vec{I}-\vec{S}^{-1})^{1/2},  \\
    \frac{1}{K-1} \anomaly{\ensLatentInno}_{K} \anomaly{\ensLatentInno}_{K} \overset{\mathrm{SVD}}{=}& \vec{U} \vec{S} \vec{U}^\T.
    \end{align}
    	\label{eq:etkfZ}
\end{subequations}

The procedure is visualised in \reffig{\ref{fig:schematic}} and summarised as \refalg{\ref{alg:double}} in \refapp{\ref{app:algorithms}}. 

\section{Experimental Setup}
\label{sec:setup}

This section starts with a description of the model and VAE used in the experiments testing different scenarios. These are followed by the results produced in the experiments.

\subsection{The dynamical model}

The \config{single} and \config{double ETKF-VAEs} are tested using a conceptual model, a discrete map that moves a point along a circle in a 2D-plane ($\modelSpace=\realSpace{2}$). The position at time $t+1$ is given by 

\begin{subequations}
\begin{align}
	x(t+1) &= x(t) \cos \alpha \psi(\vecState(t))  - y(t) \sin \alpha \psi(\vecState(t)) + \frac{x(t)}{||\vecState(t)||}A \alpha \cos(\alpha t)  \label{eq:model_x},\\
	y(t+1) &= x(t) \sin \alpha \psi(\vecState(t))  + y(t) \cos \alpha  \psi(\vecState(t)) + \frac{y(t)}{||\vecState(t)||}A \alpha \cos(\alpha t)  \label{eq:model_y},\\
	\psi(\vecState(t)) &= 2 \arctan \frac{y(t)}{x(t)+||\vecState(t)||} \: \rm{mod}\,2\pi ,\label{eq:angle} 
\end{align}
\label{eq:model}
\end{subequations}
with $\alpha=0.1$, $\omega = \frac{2\pi}{50}$, $\vecState(t)$ the position in the 2D-plane at time $t$, $x$ the position along the horizontal axis, $A$ the amplitude of the oscillation of the radius around $1$, $y$ the position along the vertical axis and $\psi(\vecState)$ the polar coordinate of $\vecState$ between $0$ and $2\pi$. Despite being simple, this model poses the EnKF with two difficulties that are exemplar of those it would face when applied to sea ice models. First, if $A=0$, the solution is constrained to a submanifold: the unit circle. Second, near $\psi \approx 0$ the ensemble has the potential to become bimodal as positions with polar coordinate $\delta \phi$, $0<\delta \phi \ll 2\pi$, are moved to $(1+\alpha)\delta \phi$ whilst those with coordinate $-\delta \phi$ are mapped to $-(1+\alpha)\delta \phi + 2 \alpha \pi$. 

\subsection{Network architecture}
\label{sec:architecture}

\begin{figure}[H]
	\centering
	\includegraphics[width=\textwidth,keepaspectratio=true]{"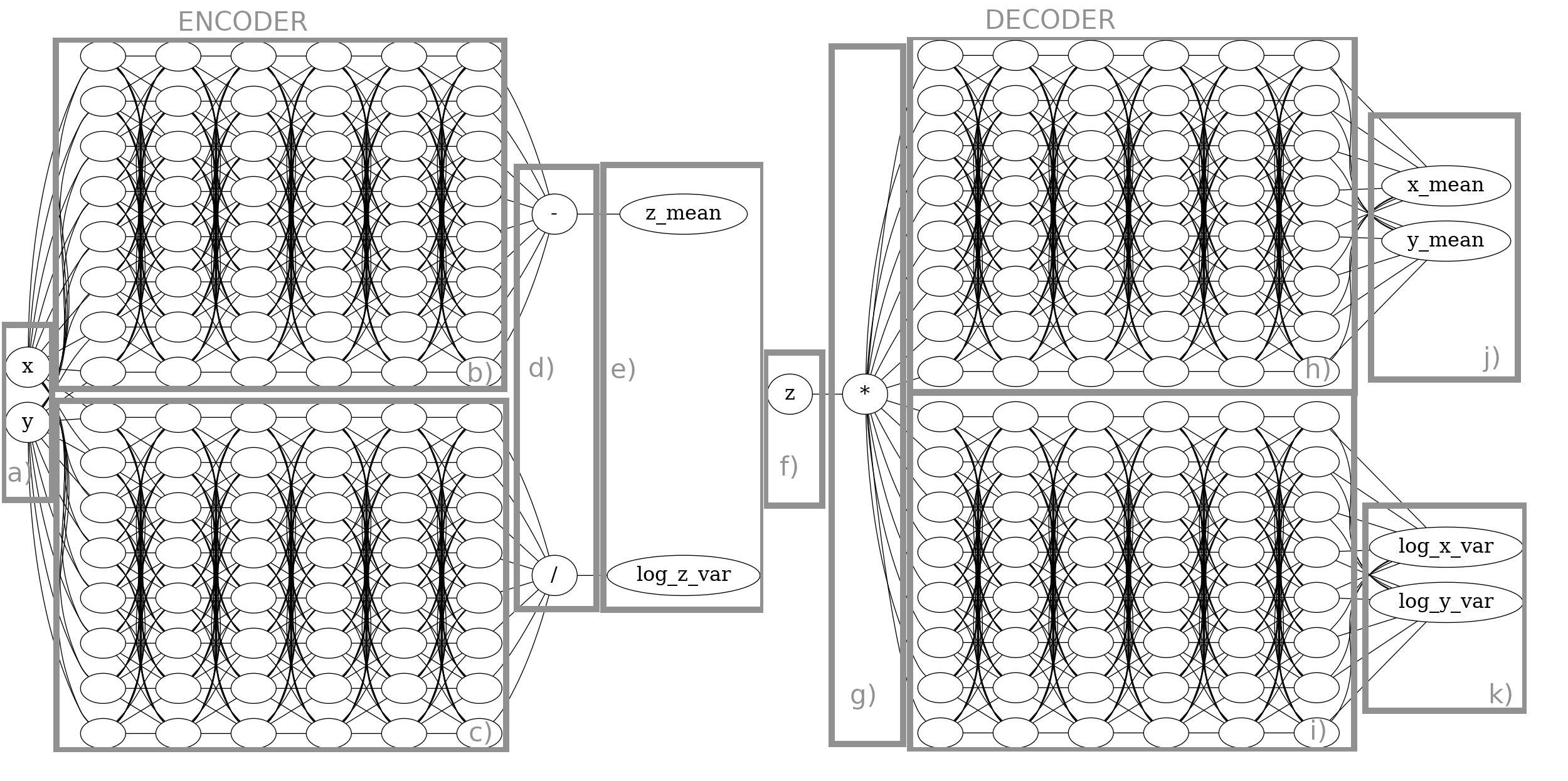"}
	\caption{Architecture of the first variational autoencoder. The encoder consists of (a) common input nodes, (b) $6$ hidden layers for the encoder predicting the mean of the conditional distribution $\vaeMean_{\encoder}$ and (c) $6$ layers for the encoder predicting its variance $\ln \vaeCov_{\encoder}$. Prior to (e) outputting, latent mean and log variance are (e) rescaled using an affine transformation (see text). The decoder (f) accepts a latent state as input, (g) applies the inverse of the latent affine transformation to the value and feeds the value to (h,i) a pair of $6$ hidden layers outputting the (j) mean $\vaeMean_{\decoder}$ and (k) log of variance $\ln \vaeCov_{\decoder}$ in state space.  For clarity only $8$ of the $32$ nodes in each hidden layer are shown.  \label{fig:architecture}}
\end{figure}

The means and logarithms of variance appearing in \refeq{\ref{eq:sample}}-\refeq{\ref{eq:kl}} are produced by multilayer perceptron. The architecture of this type of network is depicted in \reffig{\ref{fig:architecture}}. The encoder network consists of a single input layer accepting $\vecState$ and two sequences of $6$ fully connected hidden layers with $32$ nodes each, activations layers, an output layer and rescaling layer. One of the $6$-layer sequences renders $\vaeMean_{\encoder}$, the other $\vaeCov_{\encoder}$. The activation layers are Leaky Rectified Linear Units with a leakage factor of $0.1$ and the diagonal elements $\vaeCov'_{\mathrm{def}}$ in \refeq{\ref{eq:regularization}} set equal to $0.05^2$. The latent space is chosen to be one-dimensional. The rescaling layer represents an affine transformation $\for{\vecLatentState} \to a \for{\vecLatentState}+b$. Prior to the training of weights $\encoder_1$ and $\decoder_1$ $\for{\vecState}_{1},\ldots,\for{\vecState}_{\ensSize}$ are fed through the decoder and $a$, $b$ are chosen such that $\for{\vecLatentState}_{1},\ldots,\for{\vecLatentState}_{\ensSize}$ have a sample mean of $0$ and sample variance of $1$. This rescaling layer is applied in an attempt to speed up this maximisation setting the first two statistical moments of the distribution equal to that of the desired distribution $\Pnorm(\vec{0},\vec{1})$. After fixing $a$, $b$ the ELBO is maximised to find $\encoder_1$ and $\decoder_1$. The decoder consists of a similar pair of $6$-layer networks. One member of the pair produces $\vaeMean$ and the other $\ln \vaeCov$. Before entering the $6$-layer network the inverse affine transformation is applied to the latent state, i.e. $\ensLatentState \to \frac{1}{a}(\ensLatentState - b)$. 
The architecture of the second VAE is the same, the number of input nodes for the encoder and output nodes for each $6$-layer network of the decoder is equal to $N_{\obsSpace}=1$. 

Prior to any DA, the first VAE is trained on what will be called climatology. 
This climatology is generated by running the model in \refeq{\ref{eq:model}}a-c with $A=0$ for $10,000$ steps. Each $10$-th time step is set aside and the VAE is trained on this following a He normal procedure \citep{he_delving_2015} to initialise $\encoder_1$, $\decoder_1$ before training. 

\begin{figure}[H]
	\centering
	\includegraphics[height=.25\textheight,keepaspectratio=true]{"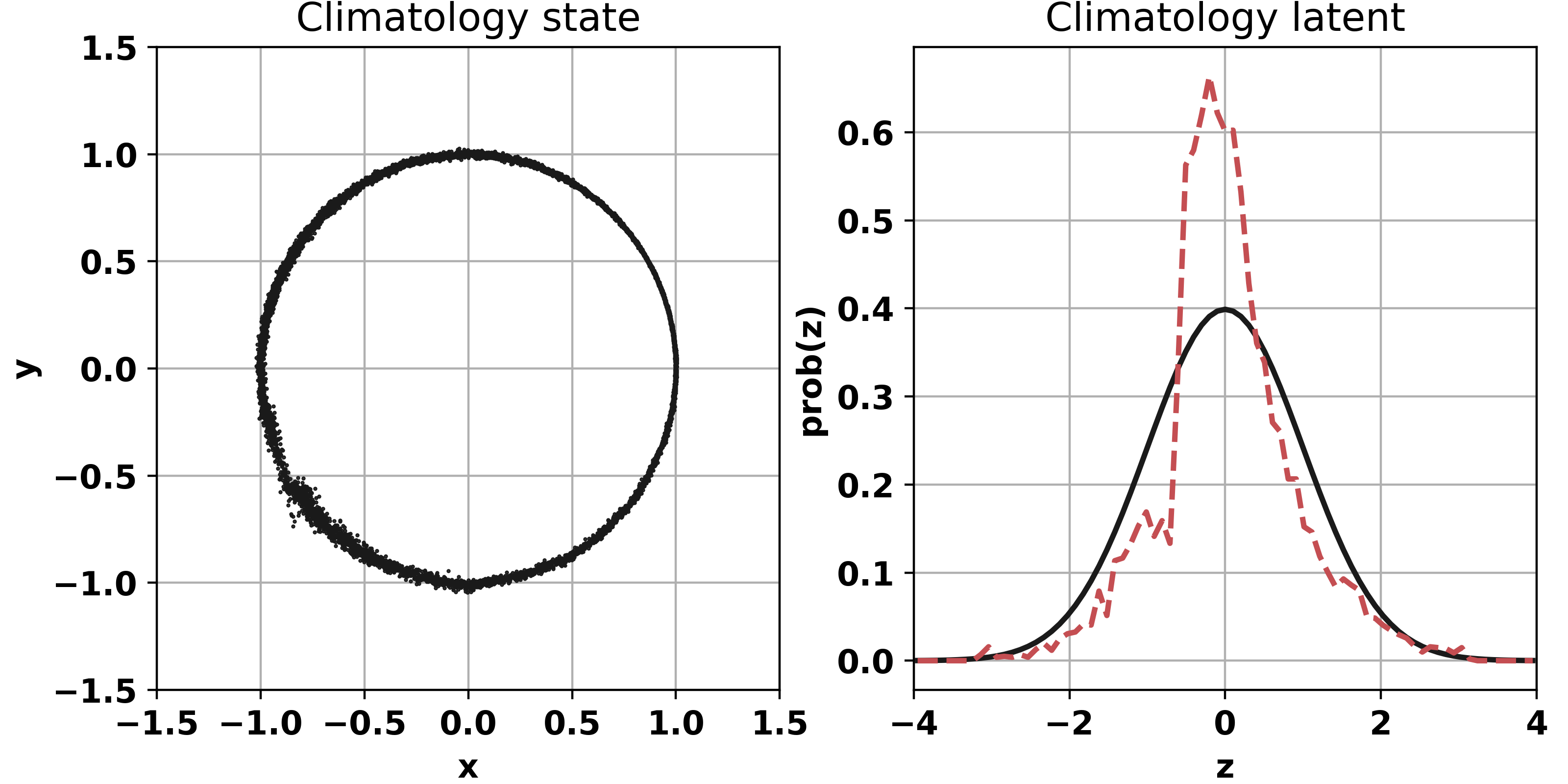"}
	\caption{(a) Climatology in the physical space generated by feeding samples from a standard normal distribution, and draw a sample from each of these using the first decoder. (b) Climatology in the latent space obtained by taking states from the climatology, and for each state drawing a sample in the latent space using the first encoder (red) together with the standard normal (black). The network architecture used $6$ dense hidden layers and $32$ nodes per hidden layer. \label{fig:climatology} } 
\end{figure}

Before settling on an architecture with $6$ hidden layers and $32$ nodes per layer, alternative layer/nodes-per-layer were tried. It was found that architectures with $6-10$ hidden dense layers and $32-64$ nodes are capable to reconstruct the unit circle. To illustrate this capability of the VAE, 1,000 samples $\vecLatentState_{i}$ were drawn from $\Pnorm(\vec{0},\vec{I})_{\latentSpace}$. For each $i$ $\vecState_{i} \sim \Pnorm(\vaeMean_{\decoder_1}(\vecLatentState_{i}),\vaeCov_{\decoder_1}(\vecLatentState_{i}))$ was drawn and added as a dot to \reffig{\ref{fig:climatology}}a. The resulting point cloud is circular, however, somewhat spread out in a band around the circle.  Next to this, a  sample $\vecLatentState_{j} \sim \Pnorm\left(\vaeMean_{\encoder_1}(\vecState_{j}),\vaeCov_{\encoder_1}(\vecState_{j})\right)$ was taken for each sample $\vecState_{j}$ from the climatology. The PDF of estimated from these samples $\vecLatentState_{j}$ is shown in \reffig{\ref{fig:climatology}}b. The figure testifies that the latent distribution approximates a standard normal, as expected based on the relation $p(\vecState)=\int p(\vecState|\vecLatentState)\Pnorm(\vecLatentState;\vec{0},\vec{I}) \, \mathrm{d}\vecLatentState$ in \refsec{\ref{sec:theory}}, but it also shows that the relation is not perfect as the estimated PDF is more strongly weighted towards $\vecLatentState=0$ than the standard normal. 

\subsection{Configurations under study}

For both the \config{single ETKF-VAE} and \config{double ETKF-VAE} architectures outlined in \refsec{\ref{sec:vae-da}} two configurations are created which differ on the type of training. In the configurations with names that contain \config{clima}, the weights of the first VAE ({\it i.e.} the VAE concerned with the model states) are copied from the VAE trained on the climatology (see \refsec{\ref{sec:architecture}}). In particular, $\encoder_1$ and $\decoder_1$ are left unchanged during the execution of the experiments. This choice, intentionally referred to as \config{clima}, mimics a general situation whereby one interrogates a dataset representing an (assumed) stationary process. As mentioned in the previous Section, the training is done using a $10,000$ time steps long trajectory, sampled every $10$ time steps. 

On the other hand, in the \config{transfer} configurations, the weights of the first VAE are retrained at each analysis step using the ensemble members current forecasts $\for{\vecState}_{1},\ldots,\for{\vecState}_{\ensSize}$ as training set. At each analysis time steps the weights, prior to training, are initialised by copying from the weights trained on the climatology. 

The second VAE is always trained at each analysis step both in the \config{clima} and \config{transfer} configurations. During the initialisation, weights are copied from the first VAE where possible, i.e. for the nodes that have the same number of connections in the first and second VAE. The remaining weights are initialised using He normalisation. An alternative approach in which weights were initialised from $\encoder_1$, $\decoder_1$ obtained during the previous analysis time step has been tried. This approach was found to be unstable as over time the weights lost the information about the shape of the circle and analysis ensemble ended up spread out through. Consequently, this approach was not pursued further.  

In addition to these configurations, a configuration without DA (\config{no DA}) and one using the standard ETKF (\config{ETKF}) have been added to facilitate comparison and benchmarking. An overview of the six configurations can be found in \reftab{\ref{tab:configs}}. 

 \begin{table}[h]
	\centering
	\normalsize\begin{tabulary}{1.0\textwidth}{CCCC}
		{\bf Configuration} & {\bf VAE mappings} & {\bf Offline training set} & {\bf Online training set} \\ 
		No DA & - & - & - \\ 
		ETKF & - & - & - \\ 
		\etkfVae{single}{clima} & $\for{\ensState} \to \for{\ensLatentState}$, $\ana{\ensLatentState} \to \ana{\ensState}$ & $\{ \clima{\vecState}(t):t\in \clima{T} \}$ & - \\ 
		\etkfVae{double}{clima}  & $\for{\ensState} \to \for{\ensLatentState}$, $\ana{\ensLatentState} \to \ana{\ensState}$,
		$\for{\ensInno} \to \for{\ensLatentInno}$ & $\{ \clima{\vecState}(t):t\in \clima{T} \}$ & $\Ima\for{\ensInno}(t)$ \\ 
		\etkfVae{single}{transfer}  & $\for{\ensState} \to \for{\ensLatentState}$, $\ana{\ensLatentState} \to \ana{\ensState}$ & $\{ \clima{\vecState}(t):t\in \clima{T} \}$ & $\Ima\for{\ensState}(t)$ \\ 
		\etkfVae{double}{transfer}  & $\for{\ensState} \to \for{\ensLatentState}$, $\ana{\ensLatentState} \to \ana{\ensState}$, $\for{\ensInno} \to \for{\ensLatentInno}$ & $\{ \clima{\vecState}(t):t\in \clima{T} \}$ & $\Ima\for{\ensState}(t)$, $\Ima\for{\ensInno}(t)$ \\ 
	\end{tabulary}
	\caption{{DA system configurations together with (1st column) states that are mapped to/from the latent space(s), (2nd column) the data set used for VAE training prior to the 1st DA step and (3rd column) prior to each DA step. Here $\Ima$ stands for image, $\ensState$ contains the ensemble members in physical space, $\ensLatentState$ the ensemble members in the latent space of the 1st VAE, $\ensInno$ the (perturbed) innovations in observation space and $\ensLatentInno$ the (perturbed) innovations in the latent space of the 2nd VAE. {\label{tab:configs}}}}
\end{table}

\section{Numerical Experiments and Results}
\label{sec:experiments}

\begin{table}[H]
	\centering
	\normalsize\begin{tabulary}{1.0\textwidth}{CCC}
		Experiment & Climate type  & Obs. error \\ 
		  I  & Stationary $(A=0)$  & Gaussian  $x \sim \Pnorm(\truth{x},0.1^2)$ \\
		II & Non-stationary $(A=0.2)$  & Gaussian $x \sim  \Pnorm(\truth{x},0.1^2)$ \\
		III & Stationary $(A=0)$  & Non-Gaussian $x \sim \Pskew(mode=\truth{x}, var=0.1^2)$ 
	\end{tabulary}
	\caption{{\label{tab:experiments}}Model setting in different experiments. Here $\Pnorm$ is the normal distribution while $\Pskew$ is the skewed normal distribution (see \refeq{\ref{eq:skewedNormal}}).}
\end{table}

Using the experimental setup in \refsec{\ref{sec:setup}} four twin experiments are carried out. In each of these experiments, $65$ models instances are run forward for $500$ time steps.  One of these model runs serves as the artificial truth, while the remaining $64$ form the ensemble. Initially, all 65 members are located on the unit circle with polar angles drawn from a uniform distribution on $[-0.1\pi, 0.1\pi]$. A summary of the different model settings and observations can be found in \reftab{\ref{tab:experiments}}. Each experiment is repeated $49$ times using $7$ different climatologies and $7$ different initial ensembles and observations per climatology. 

In \refsec{\ref{sec:fixed}} we will initially evaluate the performance of the different configurations based on correlation with, and root-mean-square difference between, the ensemble means and the truth. Evaluating DA performance by comparing the ensemble mean with the truth is standard practice for EnKFs.  This because in an EnKF the ensemble mean represents the most-likely estimate for the truth state of the system. However, this equivalence between mean and mode does not hold if the a priori distribution is non-Gaussian as is the case with the experiments in this work. Next to this, an accurate representation of the distribution itself could be of practical interest, e.g. to assess the probability that extremes occur. Therefore, metrics based on the ensemble mean might not be the best measure to evaluate the performance of the DA system. Instead, we resort to the continuous rank probability score (CRPS) as our preferred metric of performance. The CRPS is a measure for the $L_{2}$ error in the cumulative probability distribution (CDF) and is defined as 
\begin{equation}
    \mathrm{CRPS} = \int_{-\infty}^{\infty} \expectation[\left(\mathrm{CDF}(w) - \heaviside(w-\truth{w}) \right)^{2}] \,\mathrm{d}w
    \label{eq:crps}
\end{equation}
where $\heaviside$ is the Heaviside function and $w$ can be either the x-coordinate, y-coordinate, radius or angle and the integral is approximated with a numerical Lebesgue integral using the ensemble values $w^{(m)}$ for $1\leq m \leq M$ and the expectation value is calculated over all times for at which observations are available and over repetitions of the experiment. See \citet{todter_generalization_2012} for the details. 

\subsection{Stationary climate}
\label{sec:fixed}

In this experiment the truth moves along the unit circle. The system is autonomous and stationary: the circle has a constant unit radius. Every 10 time steps the x-coordinate is assimilated. Observations are drawn from the Gaussian distribution with a standard deviation of $0.1$ and centred on the x-coordinate of the truth; they are therefore unbiased.   

\begin{figure}[H]
	\centering
	\includegraphics[width=1.\textwidth,keepaspectratio=true]{"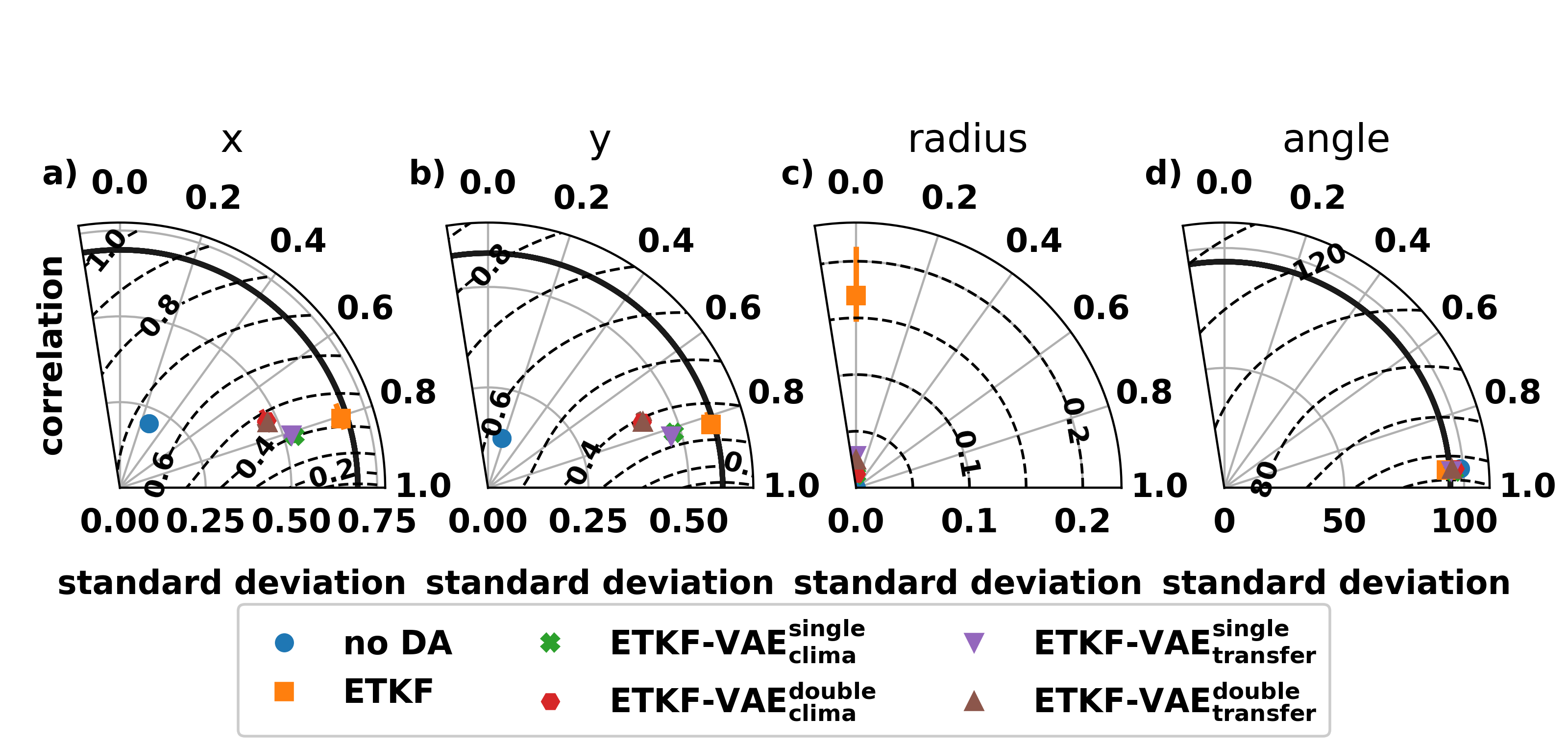"}
	\caption{Taylor diagrams for (a) x-coordinate, (b) y-coordinate, (c) radius and (d) angle. The standard deviation of the time series of the forecast mean is shown along the radial, the correlation with the truth along the azimuthal and the RMSE as dashed lines. Bars indicate the 90\%-confidence interval. \label{fig1:taylorFor}}
\end{figure}

\begin{figure}[H]
	\centering
	\includegraphics[width=1.\textwidth,keepaspectratio=true]{"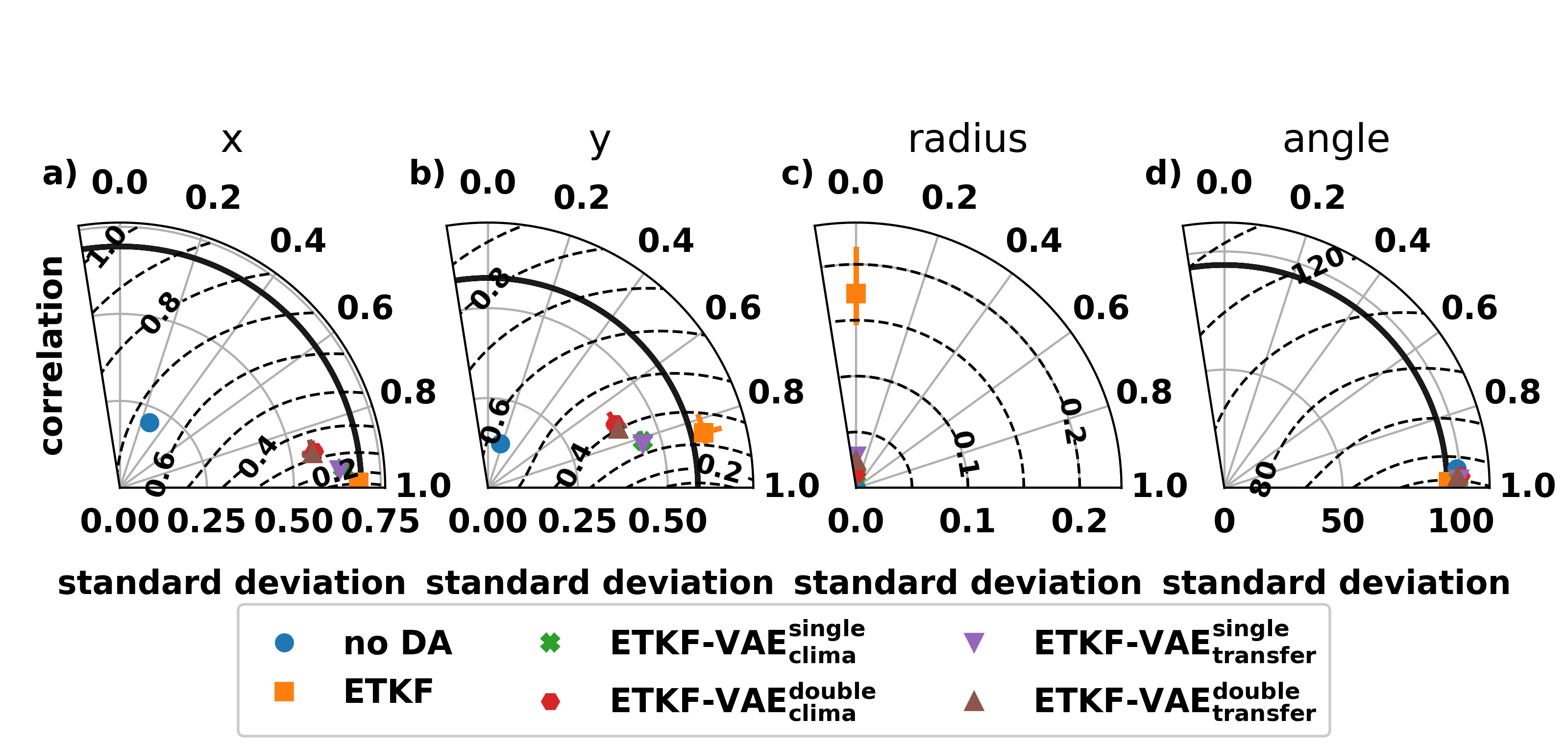"}
	\caption{As \reffig{\ref{fig1:taylorFor}} but now showing standard deviations and correlations with the mean of the analysis ensemble.  
    \label{fig1:taylorAna}}
\end{figure}

The x-coordinate, y-coordinate, radius (distance to origin), and the polar angle of the mean of the forecast/analysis ensemble are calculated just before (after) the DA correction is applied. These mean values in the different DA configurations are compared with their true values. For each repetition of the experiment, the variance of the ensemble mean, i.e. the forecast/analysis, over time, and the correlation of the ensemble mean with the truth are computed. These are then averaged over all repetitions of the experiment. For the correlations, a weighted average is used, that is, the correlation for the forecast as shown in \reffig{\ref{fig1:taylorFor}} is produced as

 \begin{equation*}
\for{\rho}=\frac{\sum_{j=1}^{999}\for{\rho}_{j} \truth{\sigma}_{j}\for{\sigma}_{j}}{\sum_{j=1}^{999} \truth{\sigma}_{j}\for{\sigma}_{j}}
\end{equation*}
with $\for{\rho}_{j}$ the correlation between the time series of a variable (x-coordinate/y-coordinate/radius/angle) obtained from the truth on one hand and the forecast mean for said variable of the $j$-th bootstrap sample on the other hand, $\truth{\sigma}_{j}$ time series standard deviation for the true value of the variable in the $j$-th bootstrap sample, and $\for{\sigma}_{j}$ the standard deviation of the forecast mean over time in the $j$-th bootstrap sample. The same procedure is followed for $\ana{\rho}$. Results are shown in \reffig{\ref{fig1:taylorFor}} and \reffig{\ref{fig1:taylorAna}}. The confidence intervals for standard deviations and correlations, as well as other metrics in the subsequent sections, have been determined using bias-corrected and accelerated bootstrapping \citep{efron_better_1987} with 999 bootstrap samples. They are shown as error bars in \reffig{\ref{fig1:taylorFor}} and \reffig{\ref{fig1:taylorAna}}. Some of them are so small that they disappear below the markers. Also shown in these diagrams, as dashed lines, is the average root-mean-squared errors (RMSEs) between the forecast/analysis means and the truth. 

\refFig{\ref{fig1:taylorFor}} shows that in both x-  and y-coordinates and in the polar angle the \config{ETKF} and the \config{ETKF-VAE} have similar RMSEs, though the \config{ETKF-VAE} configurations all slightly underestimate the variability in the x- and y-coordinates. The situation is different for the radius. Note first that for the truth and the \config{No DA} this value is constant, the standard deviation is zero and it is not possible to calculate correlations between the different configurations and the truth. In the \config{ETKF} configuration the standard deviation of the radius time series is $0.17\pm0.04$ indicating that in this experiment the reconstructed (forecasted) radius is far from being a constant unit. The \config{ETKF-VAE} configurations perform better in this regard and have standard deviations smaller than $(2.8\pm0.4)\cdot 10^{-2}$. Overall, the \config{double ETKF-VAE} configurations perform worse, in terms of the RMSE, than their \config{single ETKF-VAE} counterparts. We attribute this to the fact that sampling the innovation vector in the latent space adds an additional error. Part of this comes from the inherent stochastic nature of the encoder encapsulated in $\vaeCov_{\encoder}$ (see \refapp{\ref{app:linear_kf}} for details), part from the imperfect training of the VAE. These errors increase the spread in the latent innovation ensemble and consequently increase the covariance $\obsOpLin \for{\ensCov}\obsOpLin^{\T} +\obsCov$ as well as in the ensemble mean of the latent innovation vectors and consequently make smaller DA corrections. When we compare the forecast with the analysis metric in \reffig{\ref{fig1:taylorAna}} the performance of the different configurations is qualitatively the same. The main difference can be found in \reffig{\ref{fig1:taylorAna}}a which shows that all configurations, except \config{No DA}, exhibit higher correlations with the truth and smaller RMSEs. This is in line with expectation as the x-coordinate is directly assimilated during the experiment.  

\begin{figure}[H]       \centering\includegraphics[height=.3\textheight,width=\textwidth,keepaspectratio=true]{"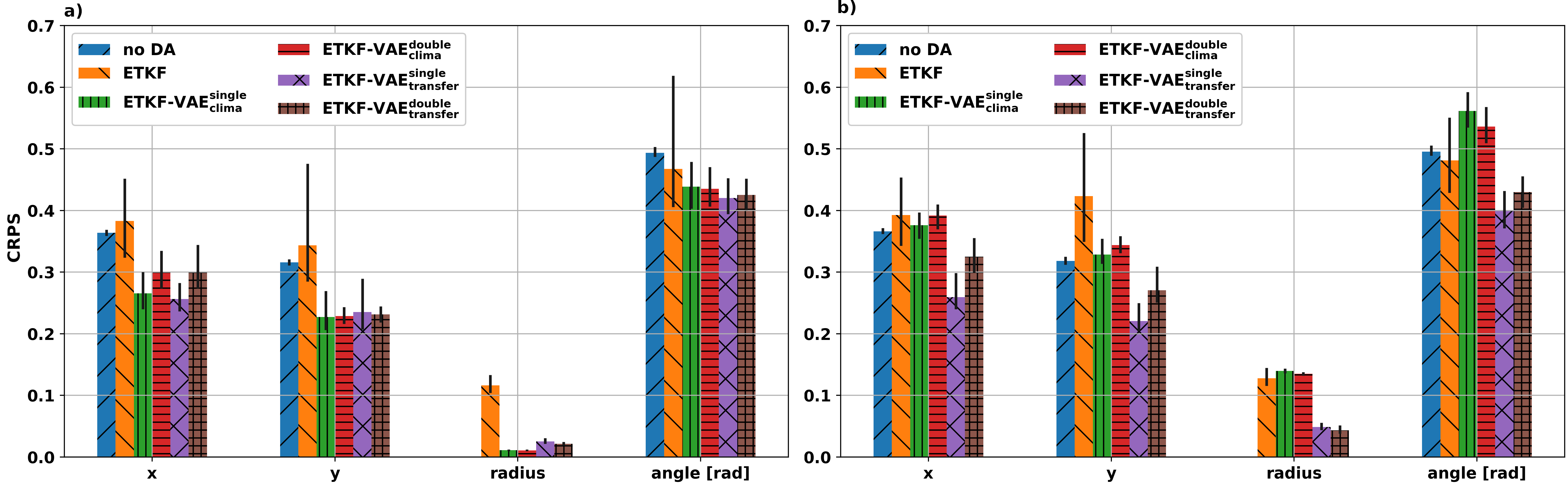"}
	\caption{CRPS of the x-coordinate, y-coordinate, radius, angle for the experiment with (a) stationary climatology and (b) with non-stationary climatology. Black lines indicate the 90\%-confidence interval. \label{fig1:crps}}
\end{figure}

The CRPS values for the different configurations and variables in this experiment are shown in \reffig{\ref{fig1:crps}a}. The CRPS for the \config{single ETKF-VAE} configurations are smaller (i.e. better) than those for the \config{ETKF} for all variables, though not significantly at the 90\%-confidence level for the angle. When using the \config{double ETKF-VAE} configurations no CRPS reduction is achieved compared to {\it ETKF} for the x-coordinate, y-coordinate and angle. Measured by the CRPS of the radius all ETKF-VAE configurations perform better than the \config{ETKF}. The reason for the improved performance for the radius is clearly visible in \reffig{\ref{fig1:crps}}a. In this figure the truth, together with the forecast and analysis ensembles are shown for time 360. In the \config{ETKF} ensemble members are moved very close to the truth, sometimes even closer than in the ETKF-VAE configurations. This reduces the RMSE effectively, but in doing so they move off the circle. On the other hand, in the ETKF-VAE configurations the analysis lies in the image of the first decoder and, if properly trained, this ensures that the analysis members end up with the correct radius close to the circle. So, the benefit of using the VAE lies mainly in its ability to restrict the analysis ensemble to the manifold of physically possible solutions. 

\subsection{Non-stationary climate}
\label{sec:varying}

In the previous section, it was shown that the use of the ETKF-VAE drastically improves the CRPS for the radius compared to \config{ETKF} as the use of a decoder for the states ensures that ensemble members stay close to the circle. This raises the question of what would happen if the submanifold containing the model solution, in our case the circle, were to change over time. This is done to mimic the effects of, for example, the change in seasons in a sea ice model or the effects of climate change in a general circulation model. To this end, we set $A=0.2$ in \refeq{\ref{eq:model}}, so that the radius of the truth will slowly vary between $0.8$ and $1.2$ in this experiment. 

Results relative to the CRPS from this non-stationary experiment are shown in \reffig{\ref{fig1:crps}b}. By comparing it with \reffig{\ref{fig1:crps}a} we see that the CRPS for x-coordinate in the \config{ETKF} is not impacted by the variation in radius of the truth, but the forecasting capacity for the non-observed y-coordinate is diminished. A larger increase in CRPS for all variables can be observed for the \config{\etkfVae{single}{clima}} and \config{\etkfVae{double}{clima}} configurations, while CRPS for the \config{\etkfVae{single}{transfer}} and \config{\etkfVae{double}{transfer}} are not significantly different at the 90\%-confidence level. We can deduce the cause of this difference by looking at \refFig{\ref{fig2:ens}}. In the \config{\etkfVae{single}{clima}} configuration (see \reffig{\ref{fig2:ens}}b), DA brings the ensemble members closer to the truth. However, the VAE in this configuration was trained on a climatology in which the radius was fixed to $1$ and it is therefore unaware of the fact that the radius of the truth changes during the model run. Consequently, the first decoder creates analysis ensemble members with a radius of $1$ instead of the radius equal to that of the truth at that specific analysis time ($1.12$). In the \config{\etkfVae{single}{transfer}} (\reffig{\ref{fig2:ens}}c) the weights of the VAE are updated at each analysis time using the forecast ensemble. This results in the analysis ensemble members being positioned at similar polar angles as in \config{\etkfVae{single}{clima}}, but now at the appropriate radius. Hence we conclude that online training of the VAE is essential if the submanifold holding the model solutions changes over time. The solution we adopted here is based on the use of the ensemble members that are already at our disposal whenever running any ensemble-based DA, thus making the method very versatile. 

\begin{figure}[H]
    \centering
	\includegraphics[height=.25\textheight,keepaspectratio=true]{"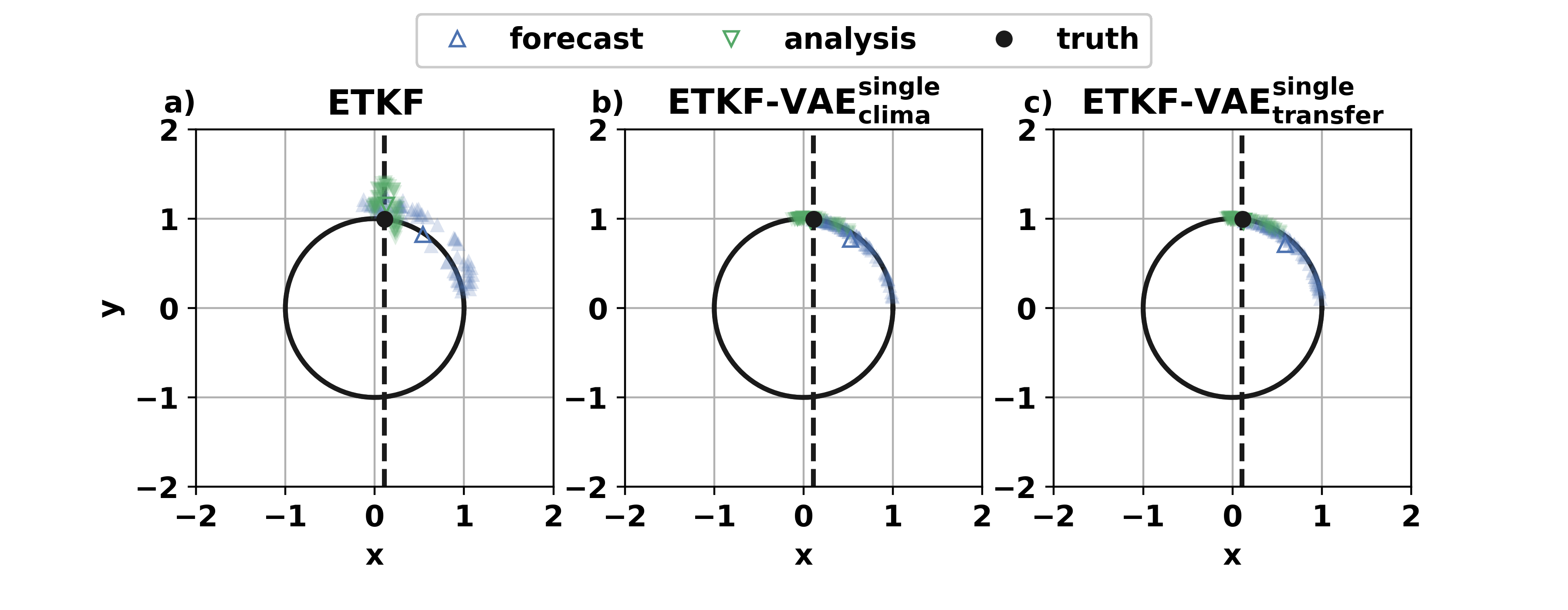"}
	\caption{(Black) truth, (blue) forecast ensemble, (green) analysis ensemble at time 360 for the (a) \config{ETKF}, (b) \config{\etkfVae{single}{clima}}, (c) \config{\etkfVae{single}{transfer}}. Ensemble means are depicted as triangles. The assimilated value of x-coordinate is depicted as a dashed black line. \label{fig1:ens}}
\end{figure}

\begin{figure}[H]
    \centering\includegraphics[height=.25\textheight,keepaspectratio=true]{"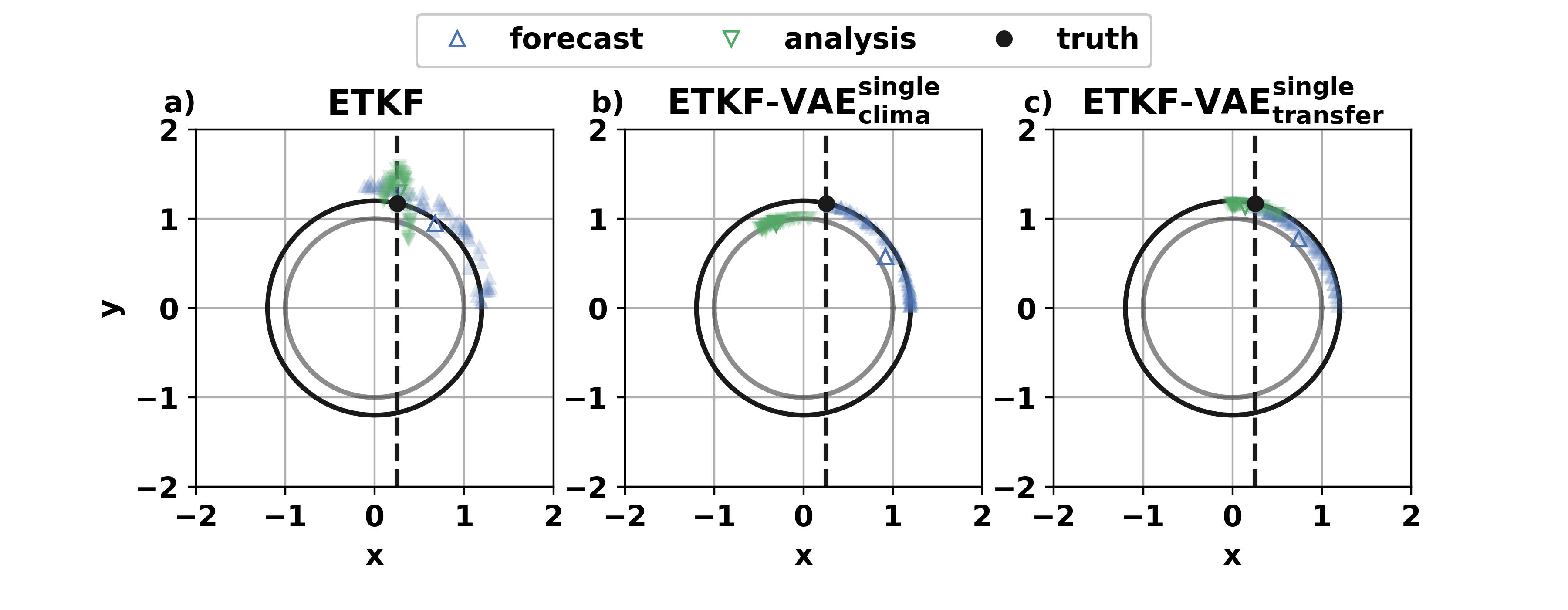"}
	\caption{(Black) truth, (blue), forecast ensemble, (green) analysis ensemble with (dashed line) observed x-coordinate at time 360 for the same configurations as in 
     \reffig{\protect\ref{fig2:ens}}. Shown are the results for the model in \refeq{\ref{eq:model}} with $A=0.2$ instead of $A=0.0$. A circle with the same radius as the truth at time 360 is depicted in black. In (b) the first VAE is trained on a climatological run in which the particle moves over the unit circle. This circle has been added in grey for reference. \label{fig2:ens}}
\end{figure}

\begin{figure}[H]
    \centering
    \includegraphics[width=0.5\linewidth, keepaspectratio]{"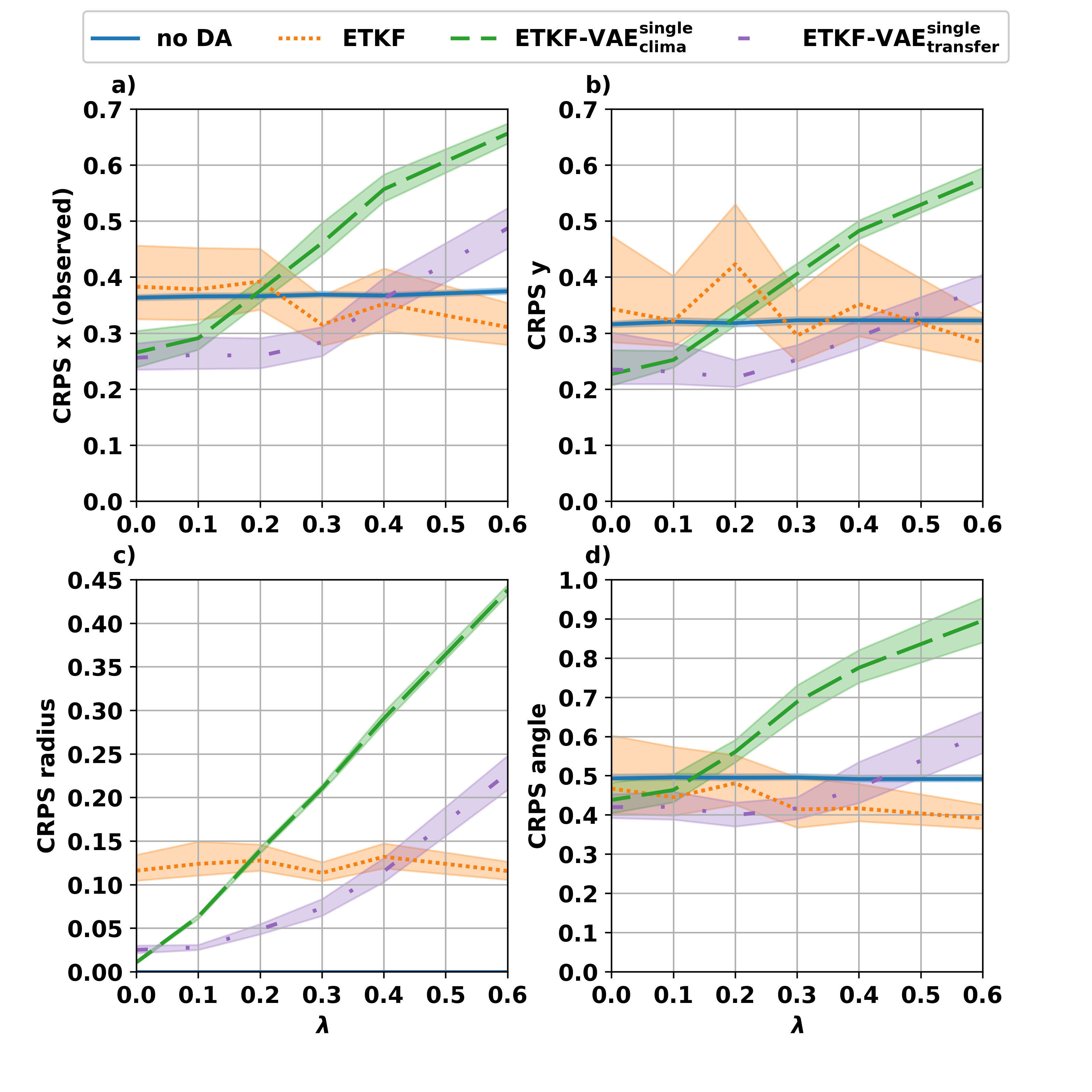"}
    \caption{CRPS for (a) the observed x-coordinate, (b) y-coordinate, (c) radius, and (d) polar angle as function of the circle radius rate of change $A$ in \refeq{\ref{eq:model}}. Results are shown for the \config{no DA}, \config{ETKF}, \config{\etkfVae{single}{clima}} and \config{\etkfVae{single}{transfer}} configurations. Bands indicate the 90\%-confidence interval of the CRPS.}
    \label{fig2:crps_amplitude}
\end{figure}

To test the robustness of our transfer learning approach, we have studied its performance against the parameter $A$ in \refeq{\ref{eq:model}} which modulates the rate at which the radius of the circle changes over time. The results for the CRPS in different variables are shown in \reffig{\ref{fig2:crps_amplitude}}. For the sake of clarity, \config{\etkfVae{double}{clima}} and \config{\etkfVae{double}{transfer}} are not shown as the ratio of their CRPSs (see \reffig{\ref{fig1:crps}b}) is qualitatively comparable to the ratio of the CRPS for \config{\etkfVae{single}{clima}} and CRPS for \config{\etkfVae{single}{transfer}}. The ETKF is unaware of the submanifold in which the truth moves and, consequently, neither of any changes in this manifold. Hence, the \config{ETKF} exhibits little dependence on $A$. The CRPS of \config{\etkfVae{single}{clima}} on the other hand increases sharply with increasing $A$ for all variables. This is consistent with the idea that as $A$ increases, the placement of the analysis ensemble member near the unit circle places them further and farther away from the truth. On the other hand \config{\etkfVae{single}{transfer}} succeeds in keeping the CRPS almost insensitive to $A$ for $A\le0.3$. It grows for $A>0.3$ but keeps its values well below the \config{\etkfVae{single}{clima}}. The deterioration of the \config{\etkfVae{single}{transfer}} skill for $A>0.3$ suggests that updating the weights of the first VAE using the ensemble fails when the actual radius of the truth is very different from the climatology. In that case, the analysis ensemble members are placed in the vicinity of the unit circle (not shown).  

\subsection{Non-Gaussian observations}
\label{sec:nongaussian}

\begin{figure}[H]
\centering
\includegraphics[height=.25\textheight,keepaspectratio]{"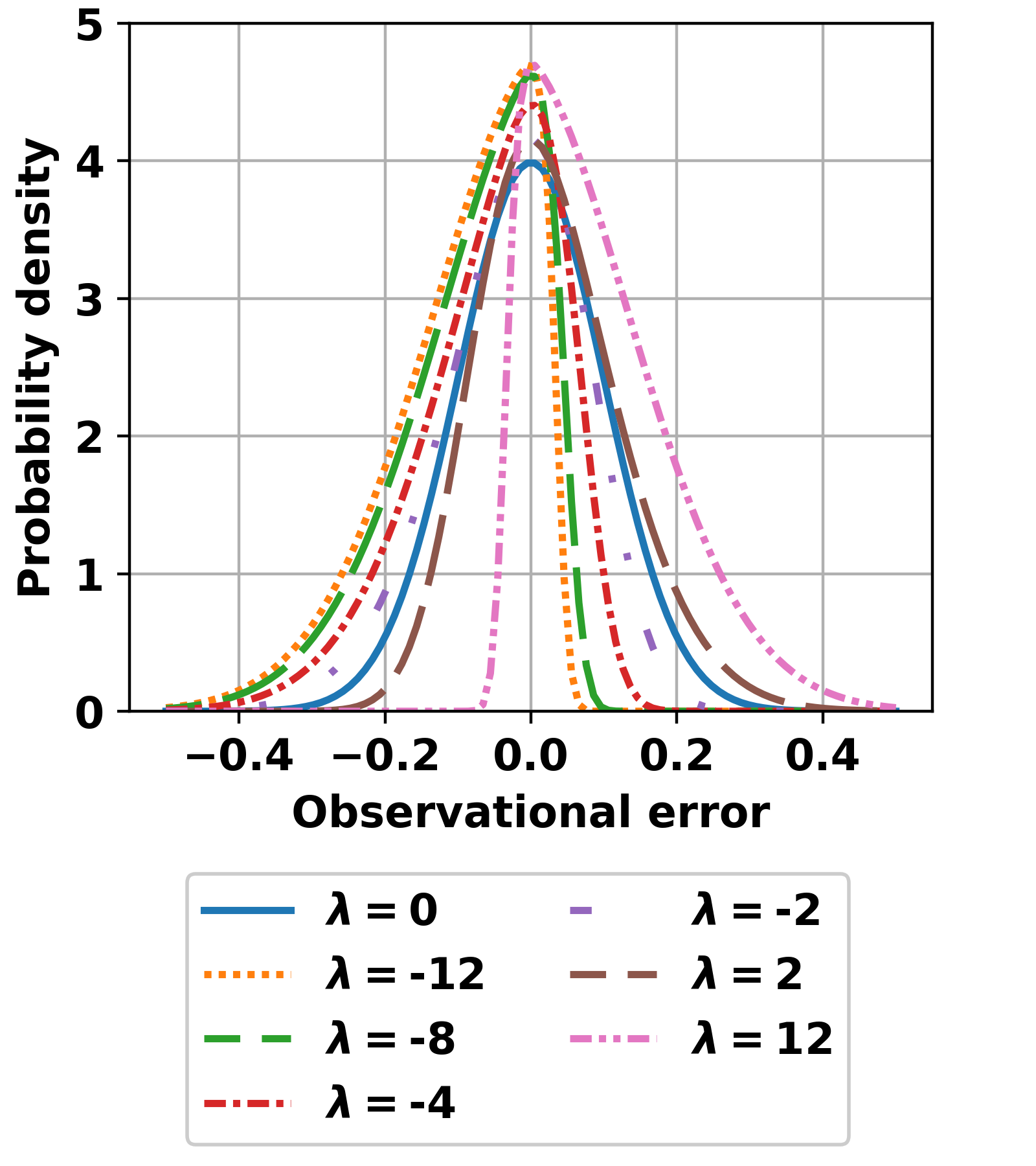"}
\caption{Skew normal probability distributions for the observational errors for different values of the skewness parameter.}
\label{fig3:skewnorm}
\end{figure}

So far non-Gaussianity was limited to the ensemble members and, indirectly, to the projection of those ensemble members into the space of observations via the observation operator. Observation errors were assumed to be Gaussian with zero-mean. In this section, we relax this assumption and instead assume that the observation errors are realisations from a skew normal distribution \citep{ohagan_bayes_1976,henze_probabilistic_1986}
\begin{equation}
\Pskew(x) = \frac{1}{\pi\sigma^{3/2}} e^{-\frac{x-\mu}{2\sigma^{2}}} 
\int_{-\infty}^{\lambda x} e^{-\frac{x-\mu}{2\sigma^{2}}} \dx,
\label{eq:skewedNormal}
\end{equation}
where $\lambda$ is the skewness parameter and $\mu$ and $\sigma$ are chosen such that the mode of the distribution is zero and the standard deviation equal to $0.1$. By doing so, the skew normal distribution reverts to the Gaussian distribution used in section~\refsec{\ref{sec:fixed}} and~\refsec{\ref{sec:varying}} if $\lambda=0$ . On the other hand, whenever $\lambda \neq 0$ the distribution \refeq{\ref{eq:skewedNormal}} is asymmetric with non-zero mean. The shape of the skew normal distribution for various values of the skewness parameter $\lambda$ is illustrated in \reffig{\ref{fig3:skewnorm}}.
The use of a skew normal violates the assumptions behind the KF, but it is not unlike a situation one encounters when assimilating, for example, satellite radiances \citep{saunders_monitoring_2013,zhu_enhanced_2014}. We are interested in exploring the capabilities of our ETKF-VAE approaches to cope with this scenario. 

\begin{figure}[H]
    \centering
    \includegraphics[width=0.5\textwidth,keepaspectratio]{"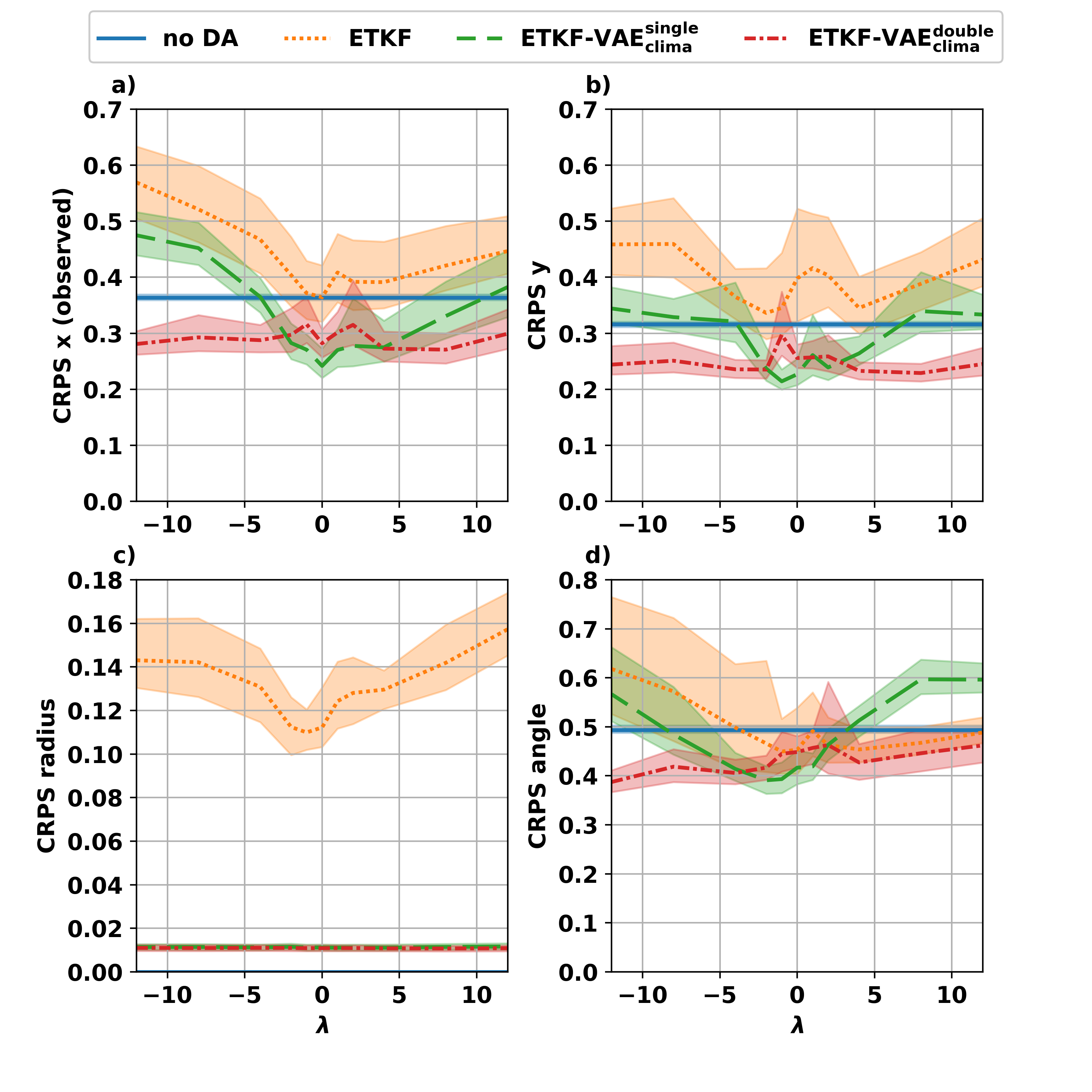"}
	\caption{CRPS for (a) the observed x-coordinate, (b) y-coordinate, (c) radius, and (d) polar angle as function of the skewness parameter $\lambda$ for the observational error distribution as appearing in \refeq{\ref{eq:skewedNormal}}. Results are shown for the \config{no DA}, \config{ETKF}, \config{\etkfVae{single}{clima}} and \config{\etkfVae{double}{clima}} configurations. Bands indicate the 90\%-confidence interval of the CRPS.} \label{fig3:crps_skew}
\end{figure}

Given that our focus here is on the impact of non-Gaussian observational errors, in this section we consider only configuration in which the training is solely on the climatology (no transfer learning) and the true circle's radius is fixed to one. The CRPS for x-coordinate, y-coordinate, radius and polar is shown in \reffig{\ref{fig3:crps_skew}} as function of the skewness parameter $\lambda$. The figure shows that the performance of \config{ETKF} in all variable deteriorates as long as the observational error distribution becomes more skewed and consequently more biased. This agrees with the findings of \citet{dee_bias_2005,lea_ocean_2008,huang_new_2020} that the EnKF performance can degenerate if observational errors are biased and no bias correction scheme is applied. \config{\etkfVae{single}{clima}} follows a similar pattern, but the impact of the bias is less pronounced. We believe that this is due to the decoder effectively limiting the extent to which the bias can impact the position of the ensemble members given that it restricts them to the circle. The situation is radically different when looking at the \config{\etkfVae{double}{clima}}. The observational error bias is contained in the synthetic innovations on which the second VAE, part of the \config{\etkfVae{double}{clima}} configuration, is trained. When the biased distribution is mapped by the encoder to a standard normal in the latent space, the bias is removed. Consequently, \config{\etkfVae{double}{clima}} does not exhibit a dependence on the skewness and outperforms \config{\etkfVae{single}{clima}} whenever the absolute value of the skewness parameter exceeds 5. It is worth noting that in this experiment the observational error distribution, and the bias in particular, is known exactly. If the distribution used to train the second VAE significantly differs from the true distribution, the risk exists that the second encoder will send the ensemble of innovations to the `wrong' part of the latent space. This may cause an incorrect rescaling of the error distribution and the benefits of using a second VAE may be small or in the worst scenario null.  

\section{Discussion \& Conclusions}
\label{sec:conclusions}

Models in the geosciences, among which the newly \nextsimdg sea ice model, generally do not satisfy the conditions underlying the KF. Background and observational errors are non-Gaussian and model and observation operators are nonlinear. EnKFs can successfully operate with when weak nonlinearity and/or small non-Gaussianity are present. In that case, however, filter updates will be suboptimal \citep{lei_comparison_2010,fowler_data_2019}, i.e. the posteriori distribution of the ensemble members does not reflect $p(\vecState|\vecObs)$.  In this work, we investigated the use of VAE to map ensemble members into realisations sampled from Gaussian distributions. This work falls under the umbrella of studies aimed at using ML to mitigate or, in the best case, to correct weaknesses of the DA procedure. We focus specifically on ensemble-based DA and, as a prototype of the latter, we choose to work with the ETKF, among the most celebrated and widely used ensemble-based approach \citep{bishop_adaptive_2001,majumdar_adaptive_2002,wei_ensemble_2006}. 
We introduced two novel formulations of the ETKF in which either one or two VAEs are merged into the ETKF's workflow. In the \config{single ETKF-VAE} the goal is to tackle the non-Gaussianity in the physical model outputs and their physical balance. The latter is identified as a phase-space submanifold to which all realisations of the model must be confined. In this work, this submanifold is the unit circle.
With the double VAE we aimed to also address the non-Gaussinity in the observational errors. 

The \config{single} and \config{double ETKF-VAE} configurations feature either an offline (climatological-like) or an online approach in which weights are retrained using transfer learning. The latter allowed us to study the impact of, and the skill against, a time varying physical submanifold for the model states. Hence, in this work  we have tested our approaches on a non-autonomous dynamics with explicit dependence on time, like one would encounter in a scenario with climate change. 

We tested these setups in a conceptual model in which a point rotates around a circle, the {\it de facto} submanifold in these experiments. We find that the main advantage of the application of the ETKF in the latent space of the VAE is that the posterior ensemble members stay close to the circle manifold, whereas the conventional ETKF places members at ``unphysical'' positions in and outside the circle. This means that the VAE is able to identify the existence of the submanifold containing the physically possible model states. The ensemble also provides a more accurate representation for the distribution of the truth. To quantify this, we have used the Continuous Rank Probability Score (CRPS). All the ETKF-VAE configurations yield better CRPSs than the standard ETKF. Our work has also shown that updating of weights for the first VAE using the ensemble members and transfer learning is essential if the manifold changes over time. Fine-tuning the VAE in this way is, in our view, a viable way of doing this. Its success in the idealistic setup employed in this study encourages its application and testing in more complex and higher-dimensional scenarios. 

The benefits of using a second VAE for the innovations, aimed at coping with non-Gaussian observational errors, are mixed. When the observational error is Gaussian or only weakly non-Gaussian, the second VAE introduces additional variability in the innovations, weakening the correlations between innovations and ensemble members and thus reducing performance compared to the offline ``clima'' configurations. On the other hand, when the non-Gaussian character of the observational error and its bias, are predominant, our findings highlight the benefit of using a second VAE for the innovations. 

Motivated by the large, and continuously growing, amount of available satellite data, compression methods that make the number of assimilated observations computationally affordable (while simultaneously maintaining high informational content) have received increased interest in recent years \citep{xu_measuring_2011,cheng_observation_2021, pasmans_tailoring_2024}. In these studies special attention was paid to how the presence of correlations in observational errors \citep{fowler_data_2019,cheng_error_2021} impacts the compression that can be achieved, that is, what percentage of observations can be removed without significantly deteriorating the DA performance. Although not pursued here among the specific goals (and made impossible by the low dimensionality of our model), we believe that in higher dimension, the second VAE will achieve an effective data compression by using a latent space of smaller dimension than the number of observations. The key advantage of our approach, in which innovations, instead of the observation \citep[see][]{cheng_multi-domain_2024}, are mapped to the latent space is that the compression will depend on both the forecast and the observational error statistics. This is important, because according to the findings of \citet{fowler_data_2019} the covariance reduction by the DA as function of the number of observations depends on both the correlation length scales in observational errors as well as the scales in the forecast errors. Extending this study to the presence of correlated observations is one of the venues left for future works.

The ensemble-size-versus-dimensionality ratio in this work is 32, much higher than the $\ll 1$ ratios typical of operational forecasting systems. Therefore, future work should also investigate how the proposed approach scales to higher-dimensional models. In particular, the ability to train the VAE weights using the ensemble members and transfer learning might falter for more realistic model setups. One possible way to mitigate this would be the inclusion of time-lagged or time-shifted ensemble members in the training data set \citep{lorenc_improving_2017} or by somewhat shifting the physical fields in space: an approach similar to the application of covariance localisation using convolution \citep{courtier_ecmwf_1998, gaspari_construction_1999, berre_filtering_2010}. This, in combination with a switch to a convolutional variational auto encoder in which the forward neural networks are replaced with convolutional networks \citep{sohn_learning_2015}, could make it feasible to retrain the VAE at each analysis time using an ensemble that is relatively small compared to the model's dimension. Alternatively, one could use model emulators, e.g. like the partial neXtSIM simulator convolutional emulator developed in \citet{durand_data-driven_2024}, to generate larger background ensembles in a computationally effective way \citep{chattopadhyay_towards_2022} or draw the ensemble members using diffusion models explicitly trained to correctly represent the background probability distribution. \citep{finn_representation_2024,li_generative_2024,price_probabilistic_2025}. 

Another point that could not be addressed in our low-dimensional setup is the tendency of VAEs to over-smooth the small scales, \citep[see, e.g.,][Fig.~2]{finn_towards_2024}. This could lead to an underestimation of the smaller scales in the analysis ensemble. For example, in a sea ice model, deformation, a process rich of small-scales in area in which the ice is damaged, might be underestimated resulting in an underestimation of the strength of the sea ice ice stresses. The straightforward mitigation fix for this effect is to increase the latent space's dimension \citep{finn_towards_2024}. 

The training of VAEs is notably highly computationally demanding. Here we found that the training in the transfer configurations can take up considerable amount of time ($\approx 0.3$ hour for a single realisation on a laptop with a NVIDIA RTX A2000) compared to running the ETKF ($\approx 1$ minute per realisation). In operational, high-dimensional, ensemble DA systems, most of the computational time is spent running ensemble members forward using the physical model. In that case, the cost of the online VAE training may be relatively smaller than in this work and thus handleable. This is another aspect of which the scalability should be checked in future work.

\section*{Acknowledgements}

This work is part of the Scale-Aware Sea Ice Project (SASIP) and is supported by grant G-24-66154 of Schmidt Sciences, LLC –– a philanthropy that propels scientific knowledge and breakthroughs towards a thriving world. CEREA is a member of Institut Pierre-Simon Laplace.

\section*{Data availability statement}

Code for the VAEs, the DA and figures can be found is embedded in the DAPPER DA framework \citep{raanes_dapper_2023} and is available as the {\it ETKF\_VAE} branch of \url{doi.org/10.5281/zenodo.7339457}. 

\appendix 

\section{ETKF-VAE algorithms}
\label{app:algorithms}

This appendix contains pseudocode running the total DA system using either \config{single ETKF-VAE} or \config{double ETKF-VAE} configurations.  he $\Puniform(a,b)$ appearing in the algorithms stands for a uniform probability distribution on the integers between $a$ up to and including $b$. 

\begin{algorithm}
\caption{Clima training}\label{alg:clima}
\begin{algorithmic}

\State Draw $\vecState \sim p_{0}(\vecState)$
\For{$t=0,1,\ldots$}
\If{$\vecObs_{t} \neq []$}
\State Add $\vecState$ as column to $\ensState$.
\EndIf
\State $\vecState \gets M_{t\to t+1}\vecState$
\EndFor

Find $\encoder_{1},\decoder_{1}=\mathrm{arg} \underset{\encoder_1,\decoder_1}{\max} \expectation_{m\sim \Puniform(1,\mathrm{rank} \ensState)}[\ELBO(\encoder_1,\decoder_1,\ensState \text{ column }m)]$ 

\end{algorithmic}
\end{algorithm}

\begin{algorithm}
\caption{Single ETKF-VAE}\label{alg:single}
\begin{algorithmic}

\State Find $\encoder_1$, $\decoder_1$ using \refalg{\ref{alg:clima}}

\For{$m=1,\ldots,\ensSize$} 
\State Draw $\for{\vecState}_{m} \sim p_{0}(\for{\vecState})$
\EndFor

\For{$t=0,1,\ldots$}

\If{$\vecObs(t) \neq []$ and \config{transfer} configuration}  

\State Find $\encoder_{1},\decoder_{1}=\mathrm{arg} \underset{\encoder_1,\decoder_1}{\max} \expectation_{m\sim \Puniform(1,\ensSize)}[\ELBO(\encoder_1,\decoder_1,\for{\vecState}_{m})]$ 
\EndIf

\If{$\vecObs(t) \neq []$} 

\For{$k=1,\ldots,K$}
\State $\ensInno_{K} \text{ column } k \gets \vecObs + \epsilon^{\vecObs}_{k}-\obsOp (\for{\vecState}_{m_k})$
\EndFor

\For{$k=1,\ldots,\ensSize$}
\State $\ensInno_{\ensSize} \text{ column } m \gets \vecObs -\obsOp (\for{\vecState}_{m})$
\EndFor

\For{$m=1,\ldots,\ensSize$} 
\State $\for{\ensLatentState} \text{ column } m \gets \sim \Pnorm(\mu_{\encoder_1}(\for{\vecState}_m),\Sigma_{\encoder_1}(\for{\vecState}_m))$
\EndFor

\State Obtain $\ana{\ensLatentState}$ from \refeq{\ref{eq:etkfD}} using $\for{\ensLatentState}$, $\ensInno_K$ and $\ensInno_{\ensSize}$. 

\For{$m=1,\ldots,\ensSize$} 
\State $\ana{\vecLatentState}_m \gets \ana{\ensLatentState} \text{ column } m$
\State Draw $\ana{\vecState}_{m} \sim \Pnorm(\mu_{\decoder_1}(\ana{\vecLatentState}_m),\Sigma_{\decoder_1}(\ana{\vecLatentState}_m))$
\EndFor

\EndIf

\For{$m=1,\ldots,\ensSize$} 
\State $\for{\vecState}_{m} \gets M_{t \to t+1}\ana{\vecState}_{m}$
\EndFor

\EndFor
\end{algorithmic}
\end{algorithm}


\begin{algorithm}
\caption{Double ETKF-VAE}\label{alg:double}
\begin{algorithmic}

\State Find $\encoder_1$, $\decoder_1$ using \refalg{\ref{alg:clima}}

\For{$m=1,\ldots,\ensSize$} 
\State Draw $\for{\vecState}_{m} \sim p_{0}(\for{\vecState})$
\EndFor

\For{$t=0,1,\ldots$}

\If{$\vecObs(t) \neq []$ and \config{transfer} configuration}  
\State Find $\encoder_{1},\decoder_{1}=\mathrm{arg} \underset{\encoder_1,\decoder_1}{\max} \expectation_{m\sim \Puniform(1,\ensSize)}[\ELBO(\encoder_1,\decoder_1,\for{\vecState}_{m})]$ 
\EndIf

\If{$\vecObs(t) \neq []$} 

\For{$k=1,\ldots,K$}
\State $\vecInno_{ij} \gets \obsOp (\for{\vecState}_{m_i}) + \epsilon^{\vecObs}_{k}-\obsOp (\for{\vecState}_{m_j})$
\EndFor
\State Find $\encoder_{2},\decoder_{2}=\mathrm{arg} \underset{\encoder_2,\decoder_2}{\max} \expectation_{i,j\sim \Puniform(1,\ensSize)}[\ELBO(\encoder_2,\decoder_2,\vecInno_{ij})]$ 

\For{$k=1,\ldots,K$}
\State $\vecInno \gets \vecObs + \epsilon^{\vecObs}_{k}-\obsOp (\for{\vecState}_{m_k})$
\State $\ensInno_{K} \text{ column } k \gets \sim\Pnorm(\mu_{\encoder_2}(\vecInno),\Sigma_{\encoder_2}(\vecInno))$
\EndFor

\For{$k=1,\ldots,\ensSize$}
\State $\vecInno \gets \vecObs -\obsOp (\for{\vecState}_{m})$
\State $\ensInno_{\ensSize} \text{ column } m \gets \sim\Pnorm(\mu_{\encoder_2}(\vecInno),\Sigma_{\encoder_2}(\vecInno))$
\EndFor

\For{$m=1,\ldots,\ensSize$} 
\State $\for{\ensLatentState} \text{ column } m \gets \sim \Pnorm(\mu_{\encoder_1}(\for{\vecState}_m),\Sigma_{\encoder_1}(\for{\vecState}_m))$
\EndFor

\State Obtain $\ana{\ensLatentState}$ from \refeq{\ref{eq:etkfD}} using $\for{\ensLatentState}$, $\ensInno_K$ and $\ensInno_{\ensSize}$. 

\For{$m=1,\ldots,\ensSize$} 
\State $\ana{\vecLatentState}_m \gets \ana{\ensLatentState} \text{ column } m$
\State Draw $\ana{\vecState}_{m} \sim \Pnorm(\mu_{\decoder_1}(\ana{\vecLatentState}_m),\Sigma_{\decoder_1}(\ana{\vecLatentState}_m))$
\EndFor

\EndIf

\For{$m=1,\ldots,\ensSize$} 
\State $\for{\vecState}_{m} \gets M_{t \to t+1}\ana{\vecState}_{m}$
\EndFor

\EndFor
\end{algorithmic}
\end{algorithm}

\newpage
\section{Kalman equations in the latent space}
\label{app:linear_kf}

In order to gain some insight in what the KF in the latent spaces of the VAEs looks like, we expand the expressions for the states in the latent space and their mean and covariance around the truth as function of the observational and forecast errors. We will use a linear approximation, i.e. terms that are second order or higher in the errors are neglected. The following derivation holds for the \config{double ETKF-VAE} configurations. However, the equivalent for the \config{single -ETKF-VAE} can be obtained by setting $\mu_{\encoder_2}(\vecInno)=\vecInno$ and $\Sigma_{\encoder_2}(\vecInno)=\vec{0}$.

For a state $\vecState \in \modelSpace$ and observation $\vecObs \in \obsSpace$ we define
\begin{subequations}
\begin{align}
    \vecLatentState =& 
    \mu_{\encoder_1}(\vecState) + \epsilon^{\vecLatentState} \approx
    \mu_{\encoder_1}(\truth{\vecState})+\Dmu{1}\epsilon^{\vecState} + \epsilon^{\vecLatentState},  \\
    \vecLatentInno =& 
    \mu_{\encoder_2}(\vecObs + \epsilon^{\vecObs} - \obsOp(\vecState)) + \epsilon^{\vecInno}
    \approx \mu_{\encoder_2}(\vecObs - \obsOp(\truth{\vecState})) - \Dmu{2}\obsOpLin \epsilon^{\vecState} + \Dmu{2} \epsilon^{\vecObs} +\epsilon^{\vecInno}, \\
    \vec{g} =& \mu_{\encoder_2}(\vecObs - \obsOp(\vecState)) \approx \mu_{\encoder_2}(\vecObs - \obsOp(\truth{\vecState})) - \Dmu{2}\obsOpLin\epsilon^{\vecState} + \epsilon^{\vecInno},
\end{align}
\end{subequations}
with $\truth{\vecState}$ the unknown true state of the model, $\obsOpLin$ the derivative of the potentially nonlinear observation operator $\obsOp$, $\epsilon^\vecObs$ the observational error, $\epsilon^\vecState$ the forecast error in state $\vecState$. $\mu_{\encoder_1}$ the function for the conditional mean in the first VAE, $\mu_{\encoder_2}$ the function for the conditional mean in the second VAE, $\Dmu{1}=\frac{\mathrm{d}\mu_{\encoder_1}}{\mathrm{d}\vecState}(\truth{\vecState})$, $\Dmu{2}=\frac{\mathrm{d}\mu_{\encoder_2}}{\mathrm{d}\vecInno}(\vecObs-\obsOp(\truth{\vecState}))$ indicating their derivatives, $\epsilon^{\vecLatentState}$ a realisation from a Gaussian with zero mean and covariance $\Sigma_{\encoder_1}(\vecState)$ and $\epsilon^\vecInno$ a realistiation from a Gaussian with zero mean and covariance $\Sigma_{\encoder_2}(\vecInno)$. 

If $\ensSize,L\gg1$ the matrix products appearing in \refeq{\ref{eq:etkfZ}} can now be approximated as
\begin{subequations}
\begin{align}
    \frac{1}{\ensSize-1}\anomaly{\for{\ensLatentState}}(\anomaly{\for{\ensLatentState}})^\T \approx& \expectation[\vecLatentState \vecLatentState^\T] = \Dmu{1}\for{\ensCov}\Dmu{1}^\T + \Sigma_{\encoder_1}, \\
    \frac{1}{L-1}\anomaly{\ensLatentInno}_{K}\anomaly{\ensLatentInno}_{K}^\T \approx& \expectation[\anomaly{\vecLatentInno}\anomaly{\vecLatentInno}^\T] = \Dmu{2} \obsCov \Dmu{2}^\T + \Dmu{2} \obsOpLin \for{\ensCov} \obsOpLin^\T \Dmu{2}^\T + \Sigma_{\encoder_2}, \\ 
    -\frac{1}{\ensSize-1}\anomaly{\ensLatentState} \anomaly{\ensLatentInno}_{\ensSize}^\T \approx& \expectation[\anomaly{\vecLatentState} (-\anomaly{\vec{g}})^\T] = \Dmu{1}\for{\ensCov}\obsOpLin^\T\Dmu{2}^\T, \\
    \frac{1}{\ensSize}\ensLatentInno_{\ensSize} \vec{1}_{\ensSize} \approx& \expectation[\vec{g}] = \mu_{\encoder_2}(\vecObs-\obsOp(\truth{\vecState})) - \Dmu{2}\obsOpLin\expectation[\epsilon^{\vecState}], \\
    \frac{1}{\ensSize} \ensLatentState \vec{1}_{\ensSize} \approx& \expectation[\vecLatentState] = \mu_{\encoder_1}(\truth{\vecState})+\Dmu{1}\expectation[\epsilon^{\vecState}].
\end{align}
\label{eq:linearized}
\end{subequations}
Here it is assumed for simplicity that observational errors are unbiased ($\expectation[\epsilon^{\vecObs}]=\vec{0}$) and, as is conventional in DA, that $\epsilon^{\vecState}$, $\epsilon^{\vecObs}$, $\epsilon^{\vecLatentState}$ and $\epsilon^{\vecInno}$ are statistically independent. 

The post-DA ensemble mean ($\ana{\mu_{\vecLatentState}}$) and covariance ($\ana{\vec{P}_{\vecLatentState}}$) in the latent space are given by $\ensState$ with $\ensLatentState$ in \refeq{\ref{eq:kf}} with $\cdot_{\vecState}$ replaced by $\cdot_{\vecLatentState}$. Substitution of $\mu_{\ensLatentState}$ and $\vec{P}_{\ensLatentState}$ with their ensemble estimates based on $\ensLatentState$, $\ensLatentInno_{\ensSize}$ and $\ensLatentInno_{K}$ as defined in \refsec{\ref{sec:double}} then gives after inserting the approximations in \refeq{\ref{eq:linearized}}a-e  
\begin{subequations}
\begin{align}
    \vec{K} \overset{\mathrm{def}}{=}& \Dmu{1}\for{\ensCov}\obsOpLin^\T \Dmu{2}^\T (\Dmu{2}\obsOpLin \for{\ensCov} \obsOpLin^\T \Dmu{2}^\T + \Dmu{2} \obsCov \Dmu{2}^\T + \Sigma_{\encoder_2})^{-1}, \\
    \ana{\mu_{\ensLatentState}} \approx& \mu_{\encoder_1}(\truth{\vecState}) + \Dmu{1}\expectation[\epsilon^{\vecState}]  + \vec{K}( \mu_{\encoder_2}(\vecObs-\obsOp(\truth{\vecState}))-\Dmu{2}\obsOpLin\expectation[\epsilon^{\vecState}]), \\ 
    \ana{\vec{P}_{\vecLatentState}} \approx& \Dmu{1} \for{\ensCov} \Dmu{1}^\T + \Sigma_{\encoder_1} - \vec{K}\obsOpLin \for{\ensCov} \Dmu{1}^\T.
\end{align}
\label{eq:latent_kf}
\end{subequations}

Based on \refeq{\ref{eq:latent_kf}}a-c, the following two observations can be made. First, if we write down the SVD of $\Dmu{1}$ and $\Dmu{2}$, e.g. $\Dmu{1}=\vec{U}_1 \vec{S}_1 \vec{V}_1^\T$, then we can see that the right singular vectors (the columns of $\vec{V}_{i}$) of $\Dmu{i}$ with $i \in \{1,2\}$ basically act as feature selectors determining which features in the state space dominate the uncertainty in the latent space. Here the singular values on the diagonal of $\vec{S}_{i}$ emphasize or de-emphasize the importance of the different features. Second, the VAEs introduce additional uncertainty in the ensemble. The additional uncertainty from the first VAE, $\Sigma_{\encoder_1}$, takes on the same role as the model error covariance in the conventional KF. The uncertainty introduced by the second VAE, $\Sigma_{\encoder_2}$, behaves as a representativeness error \citep{janjic_representation_2018}.

\bibliographystyle{agu}
\bibliography{preprint_pasmans_vae_da_20250218}

\end{document}